\documentclass[runningheads]{llncs}
\usepackage{eccv}
\usepackage{eccvabbrv}
\usepackage{graphicx}
\usepackage{booktabs}
\usepackage{multirow}
\usepackage{enumitem}
\usepackage{colortbl}
\usepackage{caption}
\usepackage{tabularx}
\usepackage{float}
\usepackage{placeins}
\usepackage[accsupp]{axessibility} 
\usepackage{algorithm}
\usepackage{algpseudocode}
\usepackage{amsmath}
\usepackage{svg}
\usepackage{bbding}
\usepackage{hyperref}
\usepackage{orcidlink}

\begin{document}

\title{Trust Your Instincts: Confidence-Driven Test-Time RL for Vision-Language-Action Models} 

\titlerunning{Trust Your Instincts: Confidence-Driven Test-Time RL for VLA Models}

% \author{Siyao Chen\inst{1} \and
% Jiakang Yuan\inst{1} \and
% Jiaxin Wang\inst{2} \and
% Tao Chen\inst{1,2}\thanks{Corresponding author.}}

% \authorrunning{Chen et al.}

% \institute{College of Future Information Technology, Fudan University, Shanghai, China\\
% \email{siyaochen25@m.fudan.edu.cn, eetchen@fudan.edu.cn} \and 
% Shanghai Innovation Institute, Shanghai, China}

\author{Siyao Chen\inst{1} \and
Jiakang Yuan\inst{1} \and
Jiaxin Wang\inst{2} \and
Tao Chen\inst{1,2,3}\thanks{Corresponding author.}}

\authorrunning{Chen et al.}

\institute{College of Future Information Technology, Fudan University, Shanghai, China\\
\email{siyaochen25@m.fudan.edu.cn, eetchen@fudan.edu.cn} \and 
Shanghai Innovation Institute, Shanghai, China \and
MoShen Intelligence, Shanghai, China}

\maketitle

\begin{abstract}

Reinforcement learning (RL) has become indispensable for pushing Vision-Language-Action Models (VLA) beyond static imitation learning. However, existing RL methods typically necessitate external environmental feedback, relying on predefined success signals to guide policy updates. 
In this work, we demonstrate that VLA models possess strong internal evaluative capabilities: in discrete-action VLAs, trajectories with higher generation confidence are significantly more likely to succeed.
Based on the observation, we introduce \textbf{T$^2$VLA} (\textbf{T}est-\textbf{t}ime \textbf{VLA}), an architecture-agnostic test-time RL framework that enables VLA models to achieve self-bootstrapping policy improvement. 
Instead of relying on external rewards, \textbf{T$^2$VLA} leverages the trajectory-level similarity to high-confidence expert demonstrations as an intrinsic reward signal. 
In addition, we propose a Confidence-Driven Dual Expert Bootstrapping mechanism. By dynamically balancing a Local Pseudo-Expert for aggressive exploration and a Global Expert Pool for training stability, \textbf{T$^2$VLA} prevents policy collapse while discovering further breakthroughs.
Extensive experiments on the LIBERO and RoboTwin benchmarks show that \textbf{T$^2$VLA} consistently outperforms supervised baselines and approaches oracle RL performance with ground-truth rewards, achieving effective improvement without external reward feedback. Furthermore, \textbf{T$^2$VLA} can adapt to distinct VLA paradigms, including both OpenVLA-OFT and the $\pi$ series.
\end{abstract}

\section{Introduction}
\label{sec:intro}

Driven by the rapid evolution of Vision-Language Models~\cite{driess2023palm, liu2023visual, karamcheti2024prismatic, peng2025chimera} (VLMs), Vision-Language-Action (VLA) models~\cite{zitkovich2023rt, kim2024openvla, black2024pi_0} have emerged as a transformative paradigm in embodied AI. Recently, Reinforcement Learning~\cite{li2025simplevla, lu2025vla, guo2025improving} (RL) has become a research hotspot for VLA, as it effectively addresses the prohibitive data collection costs and inherent generalization bottlenecks associated with traditional Supervised Fine-Tuning~\cite{brohan2022rt, mandlekar2021matters, zhao2023learning, team2024octo} (SFT). Despite their remarkable success in various embodied tasks, existing RL approaches heavily rely on external supervision signals (\textit{e.g.}, environment-provided success flags), confining their efficacy to predefined scenarios.

To break free from these predefined scenarios and unlock true autonomous self-improvement, it is crucial to empower VLA models to learn directly from their own unannotated trajectories. This transition towards the ``Era of Experience''~\cite{silver2025welcome} has recently been successfully validated in VLMs through test-time Reinforcement Learning~\cite{zuo2025ttrl} in mathematics and coding tasks, which enables models to self-optimize using purely unlabeled trajectories through majority voting. However, translating this success to VLA models exposes a critical gap: unlike math problems that yield a singular, discrete answer for easy verification, robotic manipulation requires generating continuous trajectories where multiple divergent action sequences can lead to the same goal, thereby eliminating a straightforward ground truth for self-reflection.

To bridge this gap, inspired by entropy-based confidence estimation in Large Language Models~\cite{geng2024survey, nguyen2025beyond, song2025inv}, we carefully investigate whether VLA models possess internal evaluative capabilities that can substitute for external verification. Our empirical
analysis on discrete-action VLAs (Figure~\ref{fig:motivation}) reveals a strong positive correlation between internal generation confidence and task success rates (\textit{i.e.}, trajectories with higher confidence are more likely to succeed). This observation indicates that, for token-based VLA policies, the length-normalized mean log-probability of generated trajectories can serve as a useful indicator of physical success, thereby enabling policy optimization without relying on external environmental rewards.
\begin{figure*}[t]
    \centering
    \includegraphics[width=\linewidth]{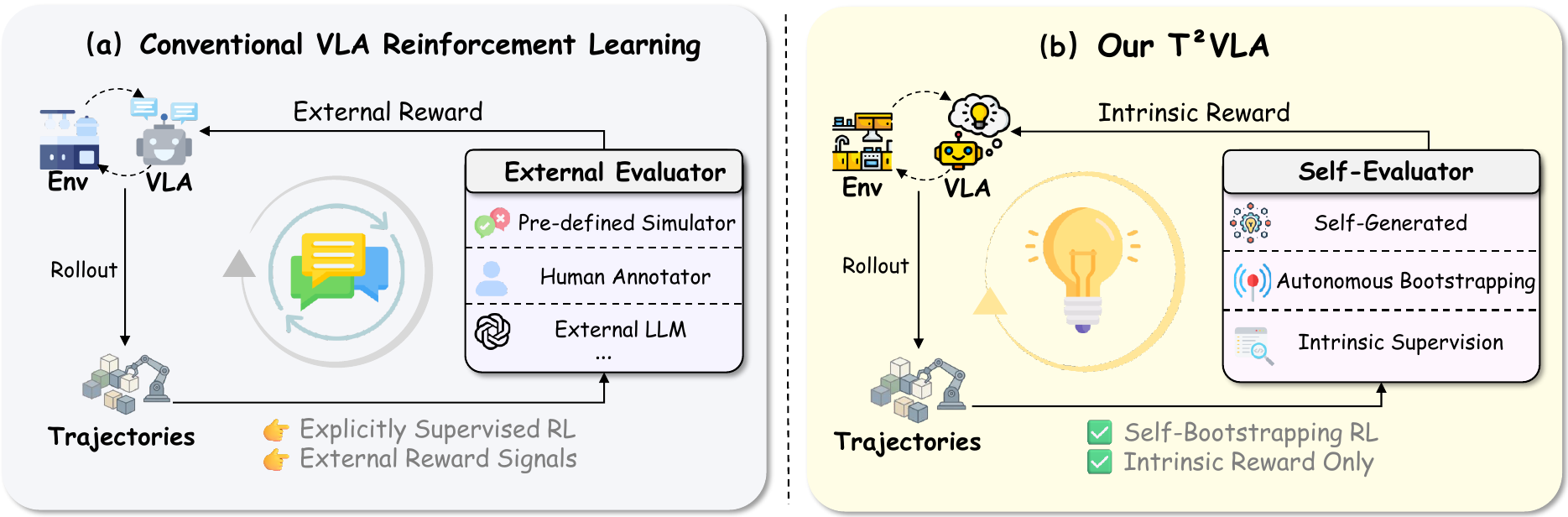}
    \caption{\textbf{Comparison of VLA Reinforcement Learning Paradigms.} \textbf{(a) Conventional VLA RL} relies on explicitly supervised RL, where policy updates are driven by external reward signals from simulators, human annotators, or external LLMs. \textbf{(b) T$^2$VLA (Ours)} introduces a self-bootstrapping RL approach. By leveraging a model-internal signal, T$^2$VLA derives rewards to drive continuous policy adaptation without external reward supervision.}
    \label{fig:paradigm}
\end{figure*}

Building upon this empirical insight, we introduce \textbf{T$^2$VLA} (\textbf{T}est-\textbf{t}ime \textbf{VLA}), an architecture-agnostic framework for self-bootstrapping policy optimization, as illustrated in Figure~\ref{fig:paradigm}. To eliminate reliance on extrinsic environmental rewards, we first propose a \textbf{Confidence-Driven Dual Expert Bootstrapping} mechanism, enabling the autonomous extraction of high-quality demonstrations directly from the model's own rollouts. Leveraging generative confidence, we elect a task-conditioned \textit{Local Pseudo-Expert} to serve as an immediate optimization anchor, steering the policy toward the most promising behaviors discovered in the current iteration. Concurrently, we maintain a priority-based \textit{Global Expert Pool} containing the top-$K$ historical trajectories, which provides a stable behavioral baseline to prevent policy regression. Subsequently, we formulate a self-bootstrapping proxy reward based on behavioral alignment with these discovered experts. We employ Dynamic Time Warping (DTW)~\cite{rakthanmanon2012searching} to evaluate sequence similarity across continuous robotic actions with varying execution horizons. The final trajectory reward is computed via an adaptive weighting strategy that dynamically interpolates between local and global DTW similarities based on their relative confidence scores. Finally, the VLA policy is updated via Group Relative Policy Optimization (GRPO)~\cite{shao2024deepseekmath}, leveraging the group-normalized proxy rewards to drive continuous, self-guided learning.

Extensive experiments on multiple benchmarks including LIBERO~\cite{liu2023libero} and RoboTwin 2.0~\cite{chen2025robotwin} demonstrate that \textbf{T$^2$VLA} can serve as an architecture-agnostic framework that yields substantial gains on various embodied tasks without relying on external rewards. Besides, \textbf{T$^2$VLA} is compatible with various VLA models such as OpenVLA-OFT~\cite{kim2025fine, li2025simplevla} and $\pi_0$~\cite{black2024pi_0}.

Our main contributions can be summarized as follows:
\begin{enumerate}
    \item We propose \textbf{T$^2$VLA}, a self-bootstrapping test-time reinforcement learning framework that eliminates the requirement for external supervision signals. By leveraging the intrinsic correlation between VLA generation confidence and physical execution success, our approach enables self-bootstrapping policy optimization without environmental reward signals.
    
    \item We introduce a confidence-driven dual-expert mechanism that synergizes a Local Pseudo-Expert for aggressive exploration with a Global Expert Pool for long-term stability. Coupled with a DTW-based adaptive reward, this provides a reliable soft anchor that prevents error compounding during unsupervised policy updates.
    
    \item We demonstrate the architecture-agnostic efficacy of \textbf{T$^2$VLA} across diverse benchmarks. Relying solely on internal signals, our method achieves over a \textbf{20\%} absolute gain on continuous-action and bimanual tasks, and elevates the average success rate of discrete-action models to \textbf{97.2\%}.
\end{enumerate}

\section{Related Work}
\label{sec:related_works}

\subsection{Vision-Language-Action Models}
Vision-Language-Action (VLA) models unify perception, reasoning, and control by directly mapping multimodal visual observations and natural language instructions to physical actions~\cite{brohan2022rt, zitkovich2023rt}. Current VLAs broadly follow two architectural paradigms based on their action representations. The first paradigm, \textbf{Discrete-Action VLAs} (e.g., RT-1~\cite{brohan2022rt}, RT-2~\cite{zitkovich2023rt}, and OpenVLA~\cite{kim2024openvla}), leverages pre-trained vision-language models (VLMs) to formulate robotic control as an autoregressive token generation problem. The second paradigm, \textbf{Continuous-Action VLAs} (e.g., Octo~\cite{team2024octo}, RDT~\cite{liu2024rdt}, $\pi_0$~\cite{black2024pi_0}, $\pi_{0.5}$~\cite{intelligence2025pi_}, and GR00T~\cite{bjorck2025gr00t}), synthesizes continuous control trajectories conditioned on multimodal embeddings via diffusion models or flow matching techniques. The standard training recipe for both classes of models is grounded in large-scale Supervised Fine-Tuning (SFT) or behavioral cloning~\cite{mandlekar2021matters}.

\subsection{Reinforcement Learning for Robotic Manipulation}
Reinforcement Learning (RL) has been widely adopted to further optimize VLA policies beyond standard SFT. \textbf{Offline RL-VLA} approaches (e.g.,  GeRM~\cite{song2024germ}, Q-Transformer~\cite{chebotar2023q}, ReinboT~\cite{zhang2025reinbot}, CO-RFT~\cite{huang2025co}) extract policies strictly from static datasets using pre-annotated rewards. To enable active environmental interaction, \textbf{Online RL-VLA} frameworks optimize policies via closed-loop feedback, extensively exploring policy gradient algorithms across diverse architectures 
  (e.g., ThinkAct~\cite{huang2025thinkact},
  RIPT-VLA~\cite{tan2025interactive},
  GRAPE~\cite{zhang2024grape},
  $\pi$RL~\cite{chen2025pi_},
  SimpleVLA-RL~\cite{li2025simplevla}).
In these online methods, reward acquisition is primarily based on explicitly designed external supervision, such as sparse binary feedback from predefined simulators~\cite{li2025simplevla, chen2025pi_} or dense proxy rewards engineered via auxiliary models~\cite{zang2025rlinf}. Instead of requiring external reward signals, our proposed \textbf{T$^2$VLA} framework achieves self-bootstrapped RL by relying entirely on intrinsic rewards.

\subsection{Test-Time Learning and Adaptation}
Test-time optimization strategies for robotic manipulation broadly fall into two paradigms: learning and adaptation. \textbf{Test-Time Learning}~\cite{wang2020tent, yuan2024reg, chen2022contrastive} dynamically updates model representations or weights during inference. For example, VLS~\cite{liu2026vls} and ADPro~\cite{li2025adpro} optimize inference-time sampling via VLM-derived gradients and geometric constraints, while EVOLVE-VLA~\cite{bai2025evolve} continuously updates policy weights using learned progress estimators. Conversely, \textbf{Test-Time Adaptation} employs training-free mechanisms to guide pre-trained policies, typically through action re-ranking via external verifiers (e.g., V-GPS~\cite{nakamoto2024steering}, VLA-Pilot~\cite{li2025towards}), MCTS-based planning (e.g., VLA-Reasoner~\cite{guo2025vla}, VLAPS~\cite{neary2025improving}), or heuristic action filtering (e.g., TACO~\cite{yang2025steering}). While these pipelines predominantly depend on auxiliary external evaluators, our \textbf{T$^2$VLA} framework
performs test-time RL using intrinsic rewards derived from model-internal signals.

\section{Method}
\label{sec:method}

\begin{figure*}[t]
    \centering
    \includegraphics[width=1\textwidth]{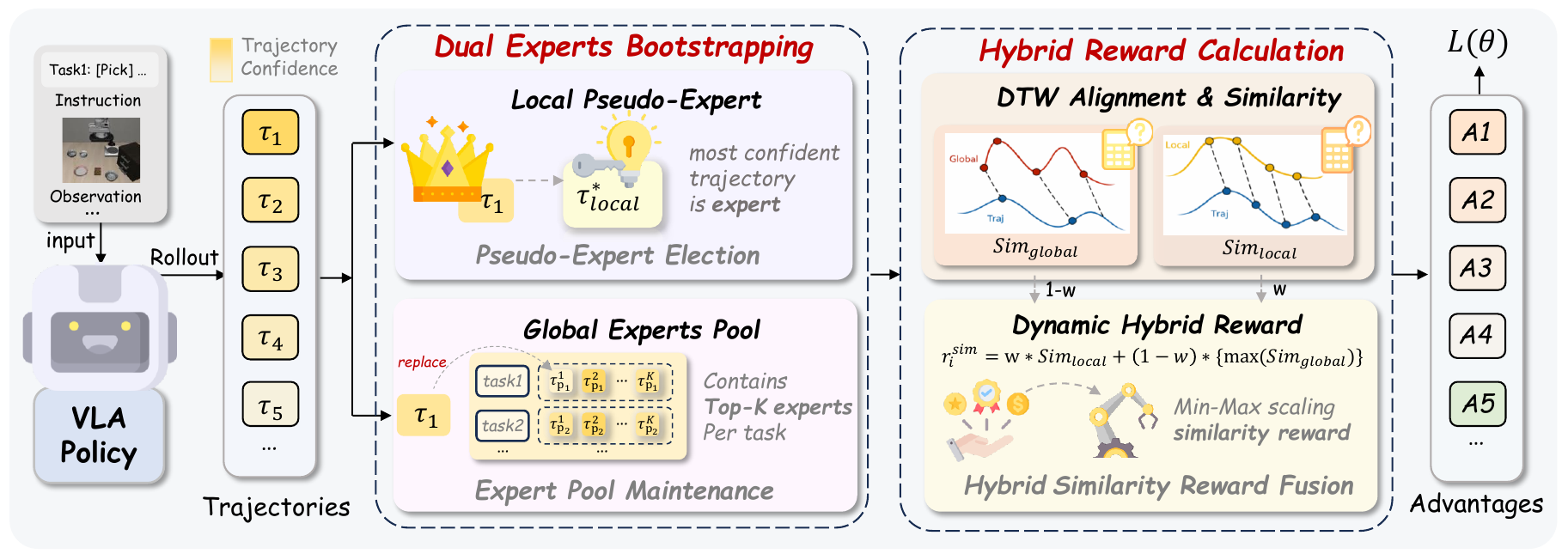} 
    \caption{\textbf{Overview of the T$^2$VLA Framework.} Our pipeline autonomously mines behavioral anchors by identifying a \textbf{Local Pseudo-Expert} from exploratory rollouts and maintaining a \textbf{Global Expert Pool}. These references are integrated via a \textbf{DTW-based Hybrid Similarity Reward} to compute advantages, enabling continuous policy optimization without external reward signals.}
    \label{fig:method_pipeline}

\end{figure*}

In this section, we present \textbf{T$^2$VLA}, a self-bootstrapping test-time RL framework for Vision-Language-Action (VLA) models (Figure~\ref{fig:method_pipeline}). Following the problem formulation and our empirical motivation (Section~\ref{subsec:preliminaries}), we introduce a confidence-driven dual-expert mechanism to autonomously mine behavioral anchors directly from exploratory rollouts (Section~\ref{subsec:expert_bootstrapping}). These anchors are subsequently translated into intrinsic signals via a DTW-based hybrid similarity reward (Section~\ref{subsec:dtw_reward}) to drive continuous policy optimization (Section~\ref{subsec:grpo}).

\subsection{Preliminaries}
\label{subsec:preliminaries}

\textbf{Problem Formulation.}
Given an embodied task, the robotic manipulation process can be formulated as a Markov Decision Process (MDP). At each timestep $t$, the VLA model observes a state $s_t = (o_t, l)$, comprising a visual observation $o_t$ and a language instruction $l$, and generates an action $a_t$ according to a parameterized policy $\pi_\theta(a_t | s_t)$. For discrete-action VLAs, the model outputs logits over discretized action tokens that are decoded into executable actions; for flow-based VLAs, actions are generated through iterative continuous refinement. 

In standard reinforcement learning for VLAs, the model interacts with the environment to collect a rollout trajectory $\tau = (s_1, a_1, \dots, s_T, a_T)$ of length $T$. Policy optimization proceeds by sampling a batch of rollouts $\{\tau_i\}_{i=1}^N$, assigning each trajectory a scalar reward $r_i$, estimating advantages, and updating $\theta$ via a policy-gradient objective. The standard formulation maximizes the expected external return:
\begin{equation}
    \max_{\theta} \; \mathbb{E}_{\tau \sim \pi_\theta} \left[ R_{\text{env}}(\tau) \right],
\end{equation}
where $R_{\text{env}}(\tau)$ denotes the external environment reward. Consequently, existing RL-based VLA optimization methods inherently rely on explicit external verification signals to guide policy updates.

\begin{figure}[t]
    \centering
    \includegraphics[width=0.95\linewidth]{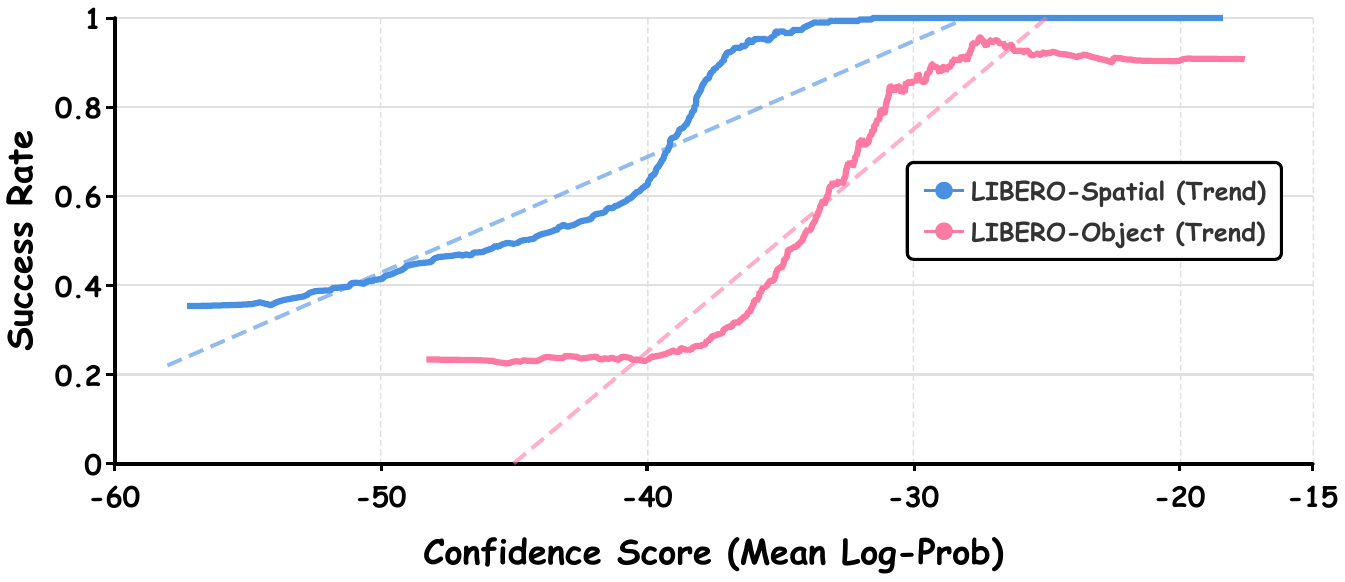} 
    \caption{\textbf{Correlation between VLA generation confidence and task success rate.} Analysis of $2,000$ OpenVLA-OFT rollouts on LIBERO suites reveals that higher mean log-probabilities consistently correlate with execution success, validating internal confidence as a reliable foundation for autonomous self-bootstrapping.}
    \label{fig:motivation}

\end{figure}

\textbf{Empirical Motivation and Reformulation.}
To eliminate the reliance on external environmental reward $R_{\text{env}}(\tau)$, we investigate whether VLA models can inherently evaluate their physical execution. As shown in Figure~\ref{fig:motivation}, empirical analysis of discrete-action VLAs reveals a strong positive correlation between trajectory-level generation confidence (defined as the mean action log-probability) and execution success. Trajectories assigned higher likelihood by the model consistently achieve higher task success rates. This observation indicates that the model's intrinsic confidence provides a robust signal regarding behavioral quality, suggesting its potential as a foundation for self-bootstrapping.

In light of this observation, we propose \textbf{T$^2$VLA}, which optimizes the VLA policy as a \textbf{self-bootstrapping reinforcement learning process} during test time. Instead of relying on external environment rewards, we replace the return with an intrinsic reward $R_{\text{self}}(\tau)$ computed solely from signals produced by the model itself. Consequently, the optimization objective transitions to:
\begin{equation}
    \max_{\theta} \; \mathbb{E}_{\tau \sim \pi_\theta} \left[ R_{\text{self}}(\tau) \right].
\end{equation}

This paradigm shift introduces a primary challenge: how to formulate an effective trajectory-level intrinsic reward $R_{\text{self}}(\tau)$ without external verification signals. While empirical evidence links confidence with task success, adopting the raw generation confidence directly as a standalone reward scalar proves inadequate for stable policy optimization. This prompts a critical question: could these high-confidence trajectories themselves serve as reliable anchors to provide effective guidance signals? To answer this, we decouple the overall challenge into two fundamental sub-problems:
(i) How to reliably extract and maintain high-quality reference trajectories from exploratory rollouts. 
(ii) How to formulate a robust trajectory reward design based on these established references.

\subsection{Confidence-Driven Dual Expert Bootstrapping}
\label{subsec:expert_bootstrapping}

To extract high-quality reference trajectories without external verification, we propose a \textbf{Confidence-Driven Dual Expert Bootstrapping} mechanism. This module is designed to autonomously mine and maintain behavioral anchors from the model's exploratory rollouts by leveraging internal confidence signals. Specifically, it consists of two primary components: the election of a task-conditioned \textit{Local Pseudo-Expert} from the current batch, and the maintenance of a dynamic \textit{Global Expert Pool} to preserve historical best practices.

 \textbf{Length-Normalized Confidence Estimation.}
  During exploration, the active policy $\pi_\theta$ generates a batch of $N$ trajectories $\mathcal{D}
  =\{\tau_1,\dots,\tau_N\}$. 
  For discrete-action VLAs, directly summing action log-probabilities inherently penalizes long-horizon
  trajectories and may favor degenerate, short-horizon behaviors. We therefore compute a length-normalized
  confidence score. For a trajectory $\tau_i$ with an effective execution horizon $T_i$, determined by
  environment termination and strictly excluding padding tokens, its confidence score is defined as the mean
  action log-probability:
  \begin{equation}
      c_i^{\mathrm{disc}}
      =
      \frac{1}{T_i}
      \sum_{t=1}^{T_i}
      \log \pi_\theta(a_{i,t}\mid s_{i,t}).
  \end{equation}
  This normalization enables confidence scores to be compared across trajectories with different execution
  horizons.

  For continuous flow-based VLAs, confidence is estimated from the Gaussian likelihood of transitions along
  the denoising process. Let $L_i$ denote the effective rollout length, measured by the number of valid
  action-chunk predictions before environment termination. At each prediction step $j$, we compute
  \begin{equation}
      \ell_{i,j}^{\mathrm{flow}}
      =
      \sum_{h,d}
      \log \mathcal{N}
      \left(
          x_{i,j,h,d}^{k+1};
          \mu_\theta(x_{i,j}^{k},k,s_{i,j})_{h,d},
          \sigma_\theta(x_{i,j}^{k},k,s_{i,j})_{h,d}
      \right),
  \end{equation}
  where $x_{i,j}^{k}$ is the intermediate denoising state at the selected denoising step $k$, while $h$ and
  $d$ index the action-chunk horizon and action dimension, respectively. The trajectory-level confidence is
  then computed as
  \begin{equation}
      c_i^{\mathrm{flow}}
      =
      \frac{1}{L_i}
      \sum_{j=1}^{L_i}
      \ell_{i,j}^{\mathrm{flow}}.
  \end{equation}

  This likelihood characterizes denoising consistency and provides an effective ranking signal for pseudo-expert election. The elected trajectories subsequently serve as behavioral references for computing intrinsic rewards. 
  
  For simplicity, we use $c_i$ to denote the corresponding trajectory-level confidence score, either
  $c_i^{\mathrm{disc}}$ or $c_i^{\mathrm{flow}}$, in the following expert election process.

\textbf{Task-Conditioned Local Expert Election.} Since generative confidence strongly correlates with execution success (Section~\ref{subsec:preliminaries}), the confidence score $c_i$ provides a reliable basis for identifying trajectories that are more likely to achieve the task. By leveraging $c_i$ to anchor the most confident executions from the exploratory rollouts, we can extract behavioral references to drive iterative policy improvement. To steer suboptimal exploratory behaviors toward these successful modes, we formally designate the most confident trajectory as a pseudo-expert.
However, during the RL training loop, a sampled batch typically encompasses trajectories across diverse environments and manipulation tasks. To ensure a fair and task-specific evaluation, we first group the trajectories based on their corresponding language instruction $l$. Within each task-specific subset $\mathcal{D}_l \subseteq \mathcal{D}$, the \textit{Local Pseudo-Expert} is explicitly elected as:
\begin{equation}
    \tau_{local, l}^* = \arg\max_{\tau_i \in \mathcal{D}_l} c_i
\end{equation}
Thus, this local expert encapsulates the most reliable behavioral mode discovered for the specific task during the current iteration, serving as a dynamic, on-policy target for immediate alignment.

\textbf{Dynamic Global Expert Pool.} Relying exclusively on the local batch expert can occasionally introduce optimization instability, particularly if an exploratory batch happens to yield overall suboptimal rollouts. To maintain training stability and preserve previously discovered high-quality behaviors, we concurrently maintain a task-conditioned dynamic \textit{Global Expert Pool}, denoted as $\mathcal{P}_l$. To ensure the high quality of this historical knowledge base, we restrict the pool's entry strictly to the elected local batch experts $\tau_{local, l}^*$, filtering out the remaining suboptimal exploratory trajectories.

At each training iteration, the newly elected local expert $\tau_{local, l}^*$ is integrated into $\mathcal{P}_l$. The pool acts as a priority-based memory buffer, retaining only the top-$K$ historical experts (with capacity $K=5$ in our implementation) sorted by their confidence scores $c$:
\begin{equation}
    \mathcal{P}_l \leftarrow \text{Top-}K \left( \mathcal{P}_l \cup \{\tau_{local, l}^*\} \right) \text{ based on } c
\end{equation}
By bounding the capacity, this mechanism evicts stale behaviors from outdated policy, ensuring the pool remains aligned with the evolving policy distribution.

\subsection{DTW-Based Hybrid Similarity Reward}
\label{subsec:dtw_reward}

To address the second sub-problem of rewarding exploratory rollouts, we design a \textbf{DTW-based Hybrid Similarity Reward}. This mechanism is designed to translate expert guidance into informative signals by aligning trajectories and balancing expertise sources. Specifically, it is composed of a DTW-based trajectory alignment to handle temporal variations, and an adaptive weighting scheme to balance current batch consensus with historical best practices.

\textbf{Trajectory Alignment via DTW.} The stochastic generative nature of VLAs frequently yields action sequences with varying temporal horizons. Standard Euclidean distance enforces strict one-to-one temporal synchronization, causing massive distance penalties under minor temporal shifts even when spatial paths are identical. As visualized in Figure~\ref{fig:dtw_vs_euclidean}, such rigid point-wise matching fails to capture true structural similarity. To decouple spatial geometry from temporal progression, we employ Dynamic Time Warping (DTW)~\cite{rakthanmanon2012searching}. By non-linearly warping the time axis to enable many-to-one state mappings, DTW aligns trajectories based on their spatial morphology rather than strict temporal indices, providing a robust similarity measure for heterogeneous rollouts.

Let $\tau_A = (a_{A,1}, \dots, a_{A,M})$ and $\tau_B = (a_{B,1}, \dots, a_{B,T})$ denote two action sequences of continuous end-effector poses $a \in \mathbb{R}^d$. To prevent dimensions with large physical magnitudes from dominating the metric, all action dimensions are independently min-max normalized to $[0, 1]$. We construct a cost matrix $D \in \mathbb{R}^{M \times T}$ via dynamic programming:
\begin{equation}
    d_{i,j} = \|a_{A,i} - a_{B,j}\|_2 + \min(d_{i-1,j}, d_{i,j-1}, d_{i-1,j-1})
\end{equation}
The cumulative distance $D_{\text{DTW}}(\tau_A, \tau_B) = d_{M,T}$ is then normalized by the maximum temporal length to ensure length-invariance and yield a bounded similarity score in $(0, 1]$:
\begin{equation}
    \text{Sim}_{\text{DTW}}(\tau_A, \tau_B) = \frac{1}{1 + \frac{D_{\text{DTW}}(\tau_A, \tau_B)}{\max(M, T)}}
\end{equation}

\textbf{Dynamic Hybrid Reward Design.} To adaptively determine the trust placed in the local batch expert, we compute the alignment of each exploratory trajectory $\tau_i$ (corresponding to task $l$) against both its task-specific local expert $\tau_{local, l}^*$ and the global pool $\mathcal{P}_l$. The dynamic hybrid similarity reward $r_i^{sim}$ incorporates the maximum similarity score among all historical experts $\tau_p \in \mathcal{P}_l$:
\begin{equation}
    r_i^{sim} = w \cdot \text{Sim}_{\text{DTW}}(\tau_i, \tau_{local, l}^*) + (1 - w) \cdot \max_{\tau_p \in \mathcal{P}_l} \text{Sim}_{\text{DTW}}(\tau_i, \tau_p)
\end{equation}
The interpolation weight $w \in [0, 1]$ balances current batch consensus and historical best practices by normalizing the local expert's confidence score $c_{local, l}^*$:
\begin{equation}
    w = \text{clip}\left( \frac{c_{local, l}^* - c_{min, l}}{c_{max, l} - c_{min, l} + \epsilon}, 0, 1 \right)
\end{equation}
where $c_{max, l}$ and $c_{min, l}$ represent the maximum and minimum confidence scores currently maintained within the task-specific pool $\mathcal{P}_l$, and $\epsilon$ ensures numerical stability. This relative measure ensures that highly confident local experts ($w \to 1$) prioritize immediate on-policy discoveries, while uncertain batches ($w \to 0$) smoothly fall back to historical experts to prevent policy degradation. A detailed analysis validating the necessity of this continuous Min-Max scaling over rigid or hard-gating alternatives is provided in Sec.~\ref{subsec:insight_analysis}.

\textbf{KL Penalty.} To prevent over-optimization and maintain the valid behavioral distribution learned during pretraining, we incorporate a Kullback-Leibler (KL) divergence penalty. The final proxy reward $r_i$ for trajectory $\tau_i$ is:
\begin{equation}
    r_i = r_i^{sim} - \beta \sum_{t=1}^{T_i} \mathbb{D}_{\text{KL}} \left( \pi_\theta(\cdot | s_{i,t}) \| \pi_{\text{ref}}(\cdot | s_{i,t}) \right)
\end{equation}
where the initial SFT model $\pi_{\text{ref}}$ acts as a behavioral anchor, and $\beta$ controls the penalty strength.

\subsection{Policy Optimization via GRPO}
\label{subsec:grpo}

Given the proxy rewards $\{r_1, \dots, r_N\}$ for the $N$ sampled trajectories, GRPO computes a baseline-free advantage $A_i$ via group normalization:
\begin{equation}
    A_i = \frac{r_i - \mu(r)}{\sigma(r) + \epsilon}
\end{equation}
where $\mu(r)$ and $\sigma(r)$ denote the batch mean and standard deviation. The policy is then updated by maximizing the clipped surrogate objective:
\begin{equation}
    \mathcal{L}(\theta) = \frac{1}{N} \sum_{i=1}^N \sum_{t=1}^{T_i} \min \left( \rho_{i,t}(\theta) A_i, \text{clip}(\rho_{i,t}(\theta), 1 - \epsilon_{clip}, 1 + \epsilon_{clip}) A_i \right)
\end{equation}
where $\rho_{i,t}(\theta) = \frac{\pi_\theta(a_{i,t} | s_{i,t})}{\pi_{\text{old}}(a_{i,t} | s_{i,t})}$ and $\epsilon_{clip}$ bounds the policy update. By iteratively maximizing $\mathcal{L}(\theta)$, the policy $\pi_\theta$ is continuously aligned with the extracted expert guidance, completing the self-bootstrapping loop.

\section{Experiments}
\label{sec:experiments}

\subsection{Experimental Setup}
\label{subsec:setup}

\textbf{Benchmarks.} We evaluate our framework on two robotic manipulation benchmarks: 
(1) \textbf{LIBERO}~\cite{liu2023libero}: We evaluate the policy's robustness and long-horizon planning capacity across four diverse task suites (Spatial, Object, Goal, and Long). 
(2) \textbf{RoboTwin 2.0}~\cite{chen2025robotwin}: We assess bimanual coordination across five tasks with varying execution horizons (100-650 steps).

\textbf{Baseline Models.} To demonstrate the architecture-agnostic applicability of \textbf{T$^2$VLA}, we instantiate our framework across two generative paradigms: 
(1) \textbf{Discrete-Action VLAs}: We adopt OpenVLA-OFT~\cite{kim2025fine}. We use the modified architecture and SFT weights provided by SimpleVLA-RL~\cite{li2025simplevla}, which optimizes the decoding head for RL compatibility. 
(2) \textbf{Continuous-Action VLAs}: We evaluate flow-matching policies $\pi_0$~\cite{black2024pi_0} and $\pi_{0.5}$~\cite{intelligence2025pi_} by adopting $\pi_{\text{RL}}$'s~\cite{chen2025pi_} SFT weights and MDP to compute confidence from denoising log-likelihoods.

\textbf{Implementation Details.} We optimize \textbf{T$^2$VLA} using GRPO~\cite{shao2024deepseekmath} with AdamW (peak LR $5 \times 10^{-6}$; cosine schedule for OpenVLA-OFT, constant for $\pi_0$/$\pi_{0.5}$). We sample $N=8$ trajectories per instruction at temperature 1.6 (OpenVLA-OFT) or 1.0 ($\pi_0$/$\pi_{0.5}$). Policy updates apply a universal initial KL penalty $\beta=0.02$. The PPO clip ratio is $[0.2, 0.28]$ for OpenVLA-OFT, and 0.2 for $\pi_0$/$\pi_{0.5}$ (with gradient clipping at 2.0 and 1.0, respectively). Execution horizons are capped at 500 steps (OpenVLA-OFT) and 480 steps ($\pi_0$/$\pi_{0.5}$, max 1024 tokens), with $\pi_{0.5}$ using 4 denoising steps.

\begin{table}[t]
\centering
\caption{\textbf{Main Results on the LIBERO Benchmark.} Success rates (\%) are reported. $\Delta$ indicates the absolute improvement over the baseline.}
\label{tab:libero_main}
\footnotesize
\resizebox{\textwidth}{!}{
\begin{tabular}{llccccc}
\toprule
\textbf{Method} & \textbf{Reward} & \textbf{Spatial} & \textbf{Object} & \textbf{Goal} & \textbf{Long} & \textbf{Avg.} \\
\midrule
\multicolumn{7}{c}{\textit{Reference: Prior SFT, RL, \& Test-Time Training Baselines}} \\
\midrule
Octo~\cite{team2024octo} & None (SFT) & 78.9 & 84.6 & 85.7 & 51.1 & 75.1 \\
OpenVLA~\cite{kim2024openvla} & None (SFT) & 84.7 & 79.2 & 88.4 & 53.7 & 76.5 \\
UniVLA~\cite{bu2025univla} & None (SFT) & 96.5 & 96.8 & 95.6 & 92.0 & 95.2 \\
VLA-RL~\cite{lu2025vla} & Env. Success & 90.2 & 94.3 & 91.8 & 82.2 & 89.6 \\
SimpleVLA-RL~\cite{li2025simplevla} & Env. Success & 99.4 & 99.1 & 99.2 & 98.5 & 99.1 \\
EVOLVE-VLA~\cite{bai2025evolve} & Learned Critic & 95.4 & 97.4 & 95.8 & 94.4 & 95.8 \\
\midrule
\multicolumn{7}{c}{\textbf{\textit{Our Framework (Self-Rewarding)}}} \\
\midrule
OpenVLA-OFT~\cite{kim2025fine, li2025simplevla} & None (SFT) & 91.6 & 95.3 & 90.6 & 86.5 & 91.0 \\
\textbf{Ours (OpenVLA-OFT)} & \textbf{Self-Reward} & \textbf{97.7} & \textbf{99.6} & \textbf{96.1} & \textbf{95.3} & \textbf{97.2} \\
\rowcolor{gray!10} \multicolumn{2}{r}{\textit{Improvement ($\Delta$)}} & \textit{+6.1} & \textit{+4.3} & \textit{+5.5} & \textit{+8.8} & \textit{+6.2} \\
\midrule
$\pi_0$ (Base)~\cite{black2024pi_0, chen2025pi_} & None (SFT) & 65.3 & 64.4 & 49.8 & 51.2 & 57.7 \\
\textbf{Ours ($\pi_0$)} & \textbf{Self-Reward} & \textbf{86.3} & \textbf{91.0} & \textbf{82.0} & \textbf{68.0} & \textbf{81.9} \\
\rowcolor{gray!10} \multicolumn{2}{r}{\textit{Improvement ($\Delta$)}} & \textit{+21.0} & \textit{+26.6} & \textit{+32.2} & \textit{+16.8} & \textit{+24.2} \\
\midrule
$\pi_{0.5}$ (Base)~\cite{intelligence2025pi_, chen2025pi_} & None (SFT) & 84.6 & 95.4 & 84.6 & 43.9 & 77.1 \\
\textbf{Ours ($\pi_{0.5}$)} & \textbf{Self-Reward} & \textbf{94.9} & \textbf{98.4} & \textbf{91.8} & \textbf{55.1} & \textbf{85.1} \\
\rowcolor{gray!10} \multicolumn{2}{r}{\textit{Improvement ($\Delta$)}} & \textit{+10.3} & \textit{+3.0} & \textit{+7.2} & \textit{+11.2} & \textit{+8.0} \\
\bottomrule
\end{tabular}
}
\end{table}

\subsection{Main Results}
\label{subsec:main_results}

In this section, we present the main evaluation results on the LIBERO and RoboTwin benchmarks. To verify the effectiveness of our method, we compare \textbf{T$^2$VLA} against a diverse spectrum of baselines, including prior SFT methods (\textit{i.e.}, UniVLA~\cite{bu2025univla}), explicitly supervised RL methods (\textit{i.e.}, VLA-RL~\cite{lu2025vla} and SimpleVLA-RL~\cite{li2025simplevla}), and test-time training (TTT) approaches (\textit{i.e.}, EVOLVE-VLA~\cite{bai2025evolve}). Further, to demonstrate the advantage of our self-bootstrapping mechanism, we directly compare the performance of our method against its respective base policies under standard data regimes: OpenVLA-OFT is fine-tuned with full demonstrations (Traj-all), while both $\pi_0$ and $\pi_{0.5}$ utilize few-shot initialization. Table~\ref{tab:libero_main} details our absolute performance gains on LIBERO, categorized by reliance on external rewards. Additionally, Table~\ref{tab:robotwin_main} reports the framework's effectiveness in bimanual control scenarios across varying execution horizons.

\textbf{Results on the LIBERO Benchmark.} 
Table~\ref{tab:libero_main} demonstrates that \textbf{T$^2$VLA} (based on OpenVLA-OFT) delivers consistent gains across all four task suites, improving upon the OpenVLA-OFT baseline by +6.2\% on average and outperforming the state-of-the-art SFT model, UniVLA~\cite{bu2025univla} (97.2\% vs. 95.2\%). Notably, despite operating entirely without external reward models or environmental supervision, \textbf{T$^2$VLA} surpasses the concurrent EVOLVE-VLA~\cite{bai2025evolve} (95.8\%), which relies on a separate foundation critic. Moreover, it substantially reduces the performance gap to the oracle-guided SimpleVLA-RL~\cite{li2025simplevla} (99.1\%), which has access to true environmental feedback. These results further demonstrate that intrinsic generative log-probabilities can provide a reliable bootstrapping signal that can boost the performance on different embodied tasks.

In addition, to verify the broad applicability of \textbf{T$^2$VLA}, we extend our evaluation to continuous flow-matching models. As detailed in Table~\ref{tab:libero_main}, our approach yields a substantial 24.2\% absolute improvement in the average success rate of $\pi_0$~\cite{black2024pi_0, chen2025pi_}. Additionally, it boosts the performance of $\pi_{0.5}$~\cite{intelligence2025pi_, chen2025pi_} on the Goal suite from 84.6\% to 91.8\%. These results confirm that \textbf{T$^2$VLA} is an architecture-agnostic approach that can be flexibly combined with various VLA models.

\begin{table}[t]
\centering
\small 
\caption{\textbf{Results on RoboTwin 2.0.} Success rates (\%) using OpenVLA-OFT. Tasks are categorized by execution horizon length.}
\label{tab:robotwin_main}

\begin{tabular*}{\linewidth}{@{\extracolsep{\fill}}lccc@{}}
\toprule
\textbf{Task Category \& Name} & \textbf{SFT Baseline} & \textbf{Ours} & \textbf{Improvement ($\Delta$)} \\
\midrule
\multicolumn{4}{l}{\textit{Short Horizon (100-130 Steps)}} \\
Lift Pot & 10.1 & \textbf{39.8} & \textit{+29.7} \\
Beat Hammer & 28.1 & \textbf{68.0} & \textit{+39.9} \\
\midrule
\multicolumn{4}{l}{\textit{Medium Horizon (150-230 Steps)}} \\
Place Empty Cup & 77.3 & \textbf{84.8} & \textit{+7.5} \\
\midrule
\multicolumn{4}{l}{\textit{Long \& Extra Long Horizon (280-650 Steps)}} \\
Handover Block & 33.1 & \textbf{44.1} & \textit{+11.0} \\
Stack Bowls & 40.6 & \textbf{59.0} & \textit{+18.4} \\
\midrule
\textbf{Average} & 37.8 & \textbf{59.1} & \textbf{\textit{+21.3}} \\
\bottomrule
\end{tabular*}
\end{table}

\textbf{Results on the Bimanual Scenario (RoboTwin 2.0).} 
To validate the versatility of \textbf{T$^2$VLA} in adapting to different tasks, we further conduct experiments on the RoboTwin 2.0 benchmark. As shown in Table~\ref{tab:robotwin_main}, \textbf{T$^2$VLA} consistently outperforms the SFT baseline across diverse execution horizons. Specifically, it achieves a remarkable performance leap on short-horizon tasks like \textit{Beat Hammer} (\textit{i.e.}, from 28.1\% to 68.0\%). For complex long-horizon tasks such as \textit{Stack Bowls}, it can also achieve an 18.4\% absolute improvement. This demonstrates that the DTW-based self-bootstrapping mechanism scales effectively to high-dimensional control scenarios without environment-provided rewards.

\FloatBarrier
\subsection{Ablation Studies}
\label{subsec:ablations}
To investigate the impact of our core designs, we conduct ablation studies on the LIBERO-Long benchmark using OpenVLA-OFT~\cite{kim2025fine, li2025simplevla}.

\begin{figure}[t]
    \centering
    \begin{minipage}[c]{0.48\linewidth}
        \centering
        \small
        \captionof{table}{\textbf{Dual Expert Ablation.} Success rates (\%) on LIBERO-Long. The dual-expert design achieves the best performance.}
        \label{tab:dual_expert}
        \begin{tabular}{lc}
        \toprule
        \textbf{Method} & \textbf{SR (\%)} \\
        \midrule
        Local Expert Only & 94.5 \\
        Global Expert Only & 93.0 \\
        \textbf{Dual Expert (Ours)} & \textbf{95.3} \\
        \bottomrule
        \end{tabular}
    \end{minipage}\hfill
    \begin{minipage}[c]{0.48\linewidth}
        \centering
        \includegraphics[width=\linewidth]{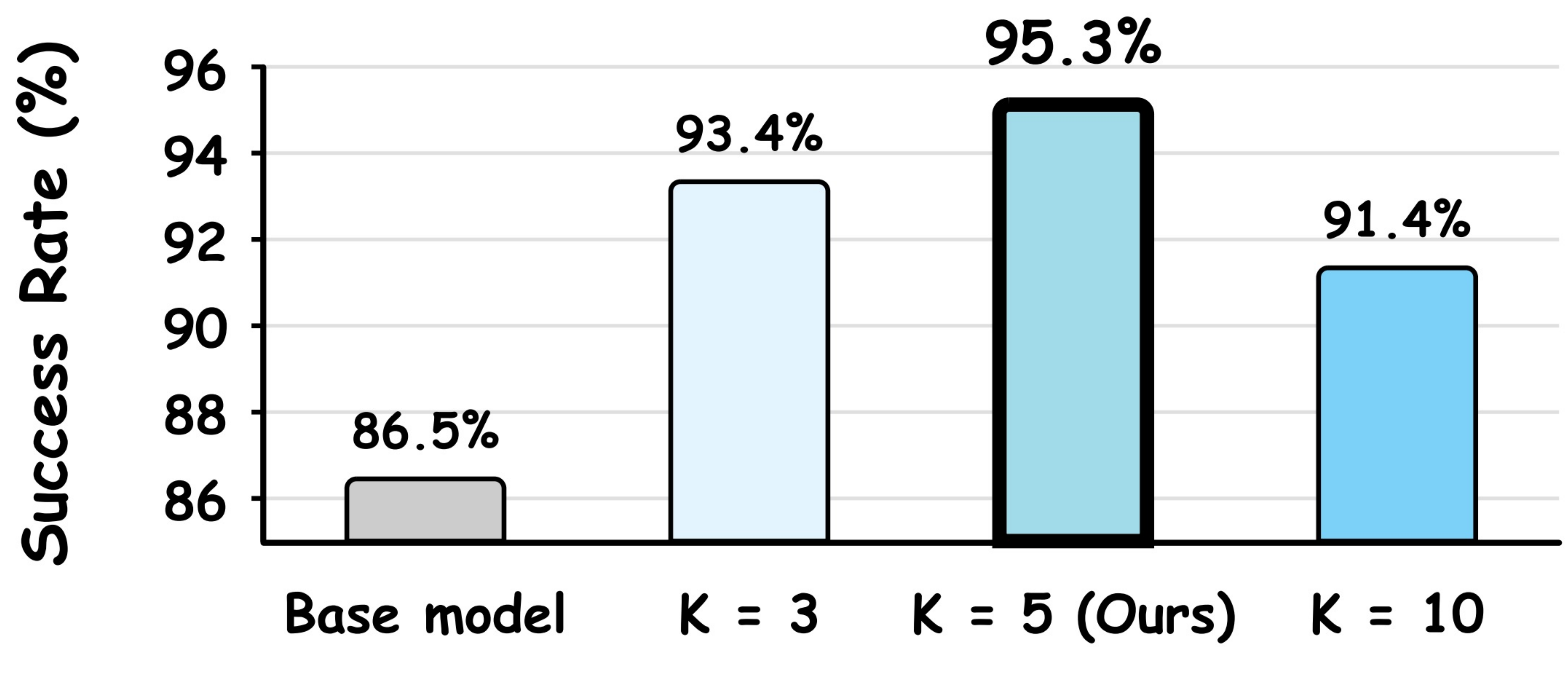}

        \caption{\textbf{Impact of Expert Pool Capacity ($K$).} Evaluated on LIBERO-Long, where $K=5$ yields the best performance.}
        \label{fig:expert_capacity}
    \end{minipage}
\end{figure}

\textbf{Effectiveness of Dual Expert Bootstrapping.} 
We first investigate the role of the proposed dual-expert bootstrapping mechanism. As detailed in Table~\ref{tab:dual_expert}, relying exclusively on the local expert captures recent high-confidence behaviors (94.5\%) but remains vulnerable to occasional suboptimal generations within a single batch. Conversely, utilizing only the global pool provides stable historical references (93.0\%) but slows down continuous adaptation, as the optimization tends to rely heavily on older trajectories. Integrating both modules effectively balances these aspects, achieving the highest success rate (95.3\%). Overall, these results indicate that the local and global experts are highly complementary: the former steers the policy toward newly discovered successful modes, while the latter serves as a stabilizing anchor to prevent policy degradation when current rollouts are poor. Their synergy yields a balanced, progressive reward signal that drives consistent performance improvements.

\textbf{Impact of Expert Pool Capacity.} 
We further analyze the sensitivity to the global expert pool capacity ($K$). As illustrated in Figure~\ref{fig:expert_capacity}, all evaluated pool capacities consistently exceed the performance of the Base model (86.5\%), validating the general effectiveness of our self-bootstrapping framework. Notably, performance peaks at a moderate capacity ($K=5$, 95.3\%), suggesting a trade-off between reference diversity and quality control. A restricted pool ($K=3$, 93.4\%) offers insufficient coverage of diverse successful behavioral modes, potentially limiting the richness of the supervision signal. On the other hand, an overly expansive pool ($K=10$, 91.4\%) retains stale trajectories from earlier, less-proficient training stages. The persistence of these outdated references can introduce suboptimal anchors into the reward calculation, which dilute the quality of the guidance signal and impede learning progress. Thus, $K=5$ provides an effective balance, maintaining sufficient historical diversity while ensuring the pool remains aligned with high-quality behavioral standards.

\FloatBarrier
\subsection{Insight Analysis}
\label{subsec:insight_analysis}

\begin{table}[t]
\centering
\small
\caption{\textbf{Comparison of Advantage Shaping Strategies.} Evaluated on LIBERO-Long. $^\dagger$ indicates severe policy collapse at early stages ($<$ 100 epochs).}
\label{tab:advantage_shaping}
\resizebox{\linewidth}{!}{
\begin{tabular}{cccc}
\toprule
\textbf{Expert Fusion Strategy} & \textbf{SR (\%)} & \textbf{Expert Fusion Strategy} & \textbf{SR (\%)} \\
\midrule
Static Weighting ($w=0.5$) & 88.3$^\dagger$ & Asymmetric Z-Score Gate & 91.4 \\
Max Routing & 87.5$^\dagger$ & Sigmoid Gate & 92.6 \\
Adaptive Margin Fallback (AMF) & 90.0 & \textbf{Min-Max Normalization (Ours)} & \textbf{95.3} \\
\bottomrule
\end{tabular}
}
\end{table}

\begin{figure}[t]
\centering
\includegraphics[width=\linewidth]{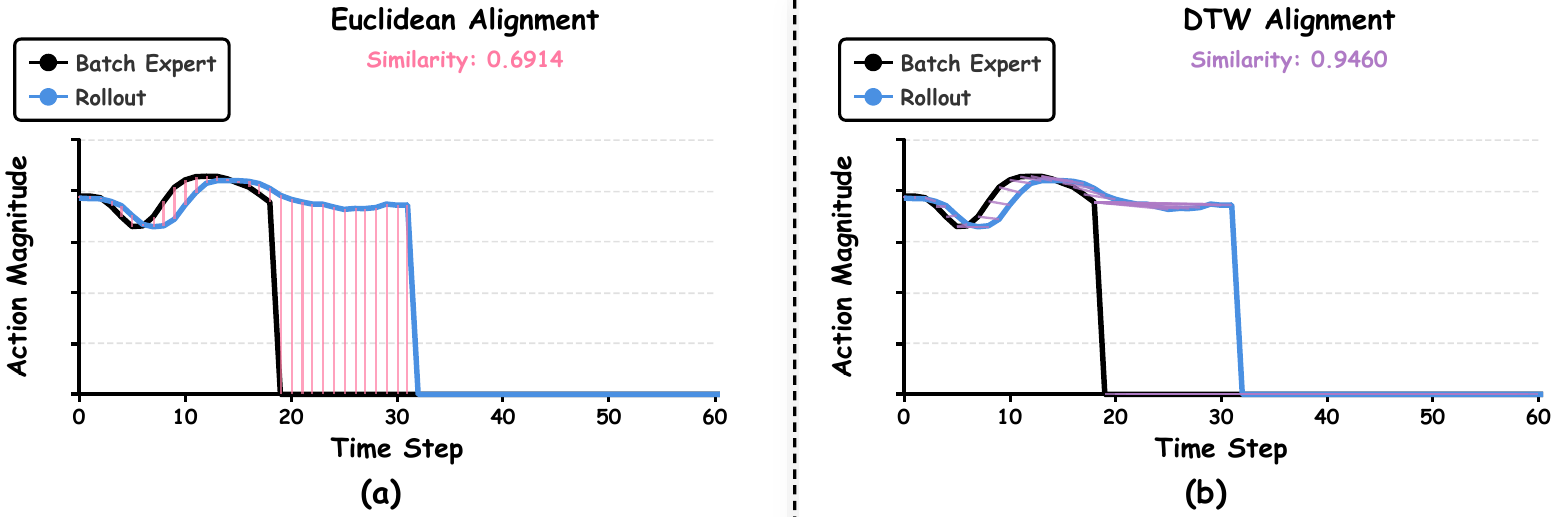}
\caption{\textbf{Trajectory alignment comparison.} In-situ training trajectories projected onto the principal action axis. (a) Rigid Euclidean matching yields high residual errors under temporal shifts. (b) DTW dynamically warps the time axis to map structurally similar states, recovering a robust spatial similarity measure.}
\label{fig:dtw_vs_euclidean}
\end{figure}

\textbf{Why Dynamic Time Warping (DTW) for Trajectory Similarity?}
We empirically justify selecting DTW over Euclidean distance for trajectory alignment. In continuous control, successful rollouts frequently exhibit temporal heterogeneities, such as pacing variations or phase shifts, compared to the reference expert. As visualized in Figure~\ref{fig:dtw_vs_euclidean}, projecting in-situ 3D trajectories onto their principal action axis reveals that Euclidean distance enforces rigid point-wise alignments. This causes residual errors to explode under minor temporal shifts, yielding an artificially depressed similarity score ($0.6914$). Conversely, DTW computes an optimal non-linear sequence alignment. By mapping corresponding geometric configurations across flexible time intervals, DTW neutralizes temporal misalignments and recovers a robust spatial congruence score ($0.9460$). This geometric-aware matching prevents high-quality exploratory rollouts from being assigned spuriously low rewards.

\textbf{The Necessity of Smooth Expert Fusion.}
Dynamically balancing the fusion weight ($w \in [0, 1]$) between the local batch and the global pool is essential for training stability. Table~\ref{tab:advantage_shaping} demonstrates that rigid assignments ($w=0.5$) or hard switching mechanisms (Max Routing) cause early policy collapse ($<$ 100 epochs) due to abrupt gradient shifts. While hard-gating approaches like Adaptive Margin Fallback (AMF) prevent this collapse, they lack fine-grained adaptation. Soft-gating strategies, such as the Temperature-scaled Sigmoid gate, offer smoother adjustments and successfully stabilize learning (92.6\%). Ultimately, our proposed Min-Max Confidence Scaling yields the best performance (95.3\%) by mapping non-stationary log-probabilities into a strictly bounded, continuous scale which ensures an adaptive and smooth integration of both experts.

\textbf{Analysis of the Bootstrapping Threshold (1-Shot Data Scarcity).}
To investigate the boundary conditions of the self-bootstrapping framework and stress-test its capability to bootstrap from minimal supervision, we evaluate OpenVLA-OFT under an extreme 1-shot data scarcity setting across all LIBERO suites. As shown in Table~\ref{tab:challenging_settings}, our method effectively improves the 1-shot SFT baseline on the Spatial, Object, and Goal suites, yielding approximately 20\% absolute gains. However, performance degrades on the Long-horizon suite (from 17.3\% to 11.0\%). This indicates a critical ``prior threshold'' for autonomous bootstrapping: for complex tasks, an excessively weak initialization causes exploratory rollouts to deviate irrecoverably from the expert demonstration. Consequently, the similarity reward degenerates into uninformative noise, and without meaningful gradient guidance, the policy optimization collapses into random drift.

\textbf{Applicability to World-Model-Generated Interactions.}
  To evaluate the applicability of \textbf{T$^2$VLA} beyond direct interaction with standard simulators, we
  conduct policy optimization using action-conditioned observations synthesized by an OpenSora world model~\cite{zheng2024open}.
  The resulting OpenVLA-OFT policy is evaluated in the original LIBERO-Spatial simulator. As shown in
  Table~\ref{tab:challenging_settings}, our method improves the success rate from 61.2\% to 63.3\%. This demonstrates that \textbf{T$^2$VLA} remains applicable when its exploratory interactions are
  generated by a learned world model.

  \begin{table}[t]
  \centering
  \small
  \caption{\textbf{Applicability under Challenging Settings.}
  Success rates (\%) under limited demonstrations and world-model-generated interactions. The degradation on the Long suite highlights the minimum prior required for successful self-bootstrapping.}
  \label{tab:challenging_settings}
  \begin{tabularx}{\linewidth}{
      @{}>{\raggedright\arraybackslash}X
      *{3}{>{\centering\arraybackslash}X}@{}
  }
  \toprule
  \textbf{Task} & \textbf{Base} & \textbf{Ours} & \textbf{$\Delta$} \\
  \midrule
  \multicolumn{4}{l}{\textit{1-Shot SFT}} \\
  LIBERO-Spatial & 63.6 & \textbf{84.0} & \textit{+20.4} \\
  LIBERO-Object  & 54.9 & \textbf{74.6} & \textit{+19.7} \\
  LIBERO-Goal    & 59.6 & \textbf{83.6} & \textit{+24.0} \\
  LIBERO-Long    & \textbf{17.3} & 11.0 & \textit{-6.3} \\
  \midrule
  \multicolumn{4}{l}{\textit{World-Model Rollouts}} \\
  OpenSora-Spatial & 61.2 & \textbf{63.3} & \textit{+2.1} \\
  \bottomrule
  \end{tabularx}
  \end{table}
  
\section{Conclusion}
\label{sec:conclusion}

In this work, we explore the self-bootstrapping optimization of Vision-Language-Action (VLA) models without relying on external rewards and propose \textbf{T$^2$VLA}, an architecture-agnostic framework for autonomous policy evolution. By autonomously mining high-quality demonstrations through the proposed confidence-driven dual expert
  bootstrapping mechanism and constructing a dynamic DTW-based hybrid similarity reward, \textbf{T$^2$VLA}
  enables VLA models to continuously refine their execution using intrinsic signals. Our \textbf{T$^2$VLA} establishes a scalable, intrinsic-reward-driven paradigm, demonstrating that a model's internal signals can effectively bootstrap continuous and autonomous policy improvement.

  \section*{Acknowledgements}
  This work is supported by National Key R\&D Program of China
  (No.~2026YFE\allowbreak0101200).
  It is also supported by Shanghai Natural Science Foundation
  (No.~23ZR\allowbreak1402900).
  The computations in this research were performed using the CFFF platform of Fudan University.

\bibliographystyle{splncs04}
\bibliography{arxiv}

\clearpage
\appendix
\section*{Appendix}

\section{Overview}
\label{sec:appendix_overview}
This appendix provides additional technical details and experimental results for \textbf{T$^2$VLA}. The content is organized as follows:
\begin{itemize}[leftmargin=*]
    \item Section~\ref{sec:benchmarks_detail}: Details of Evaluation Models and Benchmarks.
    \item Section~\ref{sec:further_confidence_analysis}: Further Analysis of Confidence-Based Expert Election.
    \item Section~\ref{sec:more_experiments}: More Experimental Results.
    \item Section~\ref{sec:appendix_shaping_math}: Mathematical Formulations of Advantage Shaping Strategies.
    \item Section~\ref{sec:algo_implementation}: Algorithmic Implementation.
    \item Section~\ref{appendix:confidence_analysis}: Analysis of Raw Confidence-based Optimization.
    \item Section~\ref{sec:learning_dynamics}: Analysis of Learning Dynamics.
    \item Section~\ref{sec:ablation_learning_dynamics}: Details of Ablation Studies on the Dual Expert Mechanism.
    \item Section~\ref{appendix:hyperparameters}: Training Hyperparameters.
    \item Section~\ref{sec:case_study_threshold}: Case Study: Visualizing the Bootstrapping Threshold.
    \item Section~\ref{sec:limitations}: Limitations.
\end{itemize}

\section{Details of Evaluation Models and Benchmarks}
\label{sec:benchmarks_detail}

\subsection{VLA Architectures}

In addition to the OpenVLA-OFT and $\pi_0$/$\pi_{0.5}$ policies evaluated in the main manuscript, we consider two additional VLA architectures:
\begin{itemize}[leftmargin=*]
    \item \textbf{StarVLA}~\cite{community2026starvla}: A discrete-action VLA that autoregressively generates action tokens.
    \item \textbf{GR00T}~\cite{bjorck2025gr00t}: A continuous-action VLA that generates action trajectories through a denoising process.
\end{itemize}

\subsection{Evaluation Environments}

We evaluate \textbf{T$^2$VLA} on three simulation benchmarks with distinct manipulation requirements:

\begin{itemize}[leftmargin=*]
    \item \textbf{LIBERO}~\cite{liu2023libero}: This benchmark evaluates compositional generalization and long-horizon planning in tabletop manipulation. We report results on its four task suites:
    \begin{itemize}
        \item \textit{LIBERO-Spatial}: Evaluates policy robustness against variations in initial object arrangements and spatial layouts.
        \item \textit{LIBERO-Object}: Examines the generalization of manipulation skills to different object instances within the same semantic category.
        \item \textit{LIBERO-Goal}: Assesses the capacity to achieve diverse task end-states using a consistent set of objects.
        \item \textit{LIBERO-Long}: Tests long-horizon stability in multi-step sequential tasks that are prone to compounding execution errors.
    \end{itemize}
    
    \item \textbf{RoboTwin 2.0}~\cite{chen2025robotwin}: This benchmark focuses on bimanual coordination and execution consistency in high-dimensional action spaces under domain randomization. Following the classification in SimpleVLA-RL~\cite{li2025simplevla}, tasks are categorized into four horizon levels based on the required planning steps:
    \begin{itemize}
        \item \textit{Short Horizon (112--130 steps)}: \textbf{Lift Pot} and \textbf{Beat Hammer}. These tasks represent scenarios with limited planning horizons, averaging 121 steps.
        \item \textit{Medium Horizon (151--223 steps)}: \textbf{Place Empty Cup}. This category covers tasks with intermediate planning horizons, averaging 176 steps.
        \item \textit{Long \& Extra Long Horizon (283--637 steps)}: \textbf{Handover Block} and \textbf{Stack Bowls}. These scenarios involve extended execution cycles, with step counts ranging from 283 to 637.
    \end{itemize}

    \item \textbf{RoboCasa}~\cite{nasiriany2024robocasa}: This benchmark provides large-scale household manipulation tasks in diverse kitchen scenes. We use the \textit{Close Drawer} task to evaluate whether \textbf{T$^2$VLA} transfers to a household simulation environment beyond LIBERO and RoboTwin 2.0.
\end{itemize}

\textbf{Group-Level Synchronization for Domain Randomization.} 
Notably, to ensure fair GRPO advantage computation in RoboTwin 2.0, we design a specific group-level synchronization mechanism to handle the randomized initial states (e.g., varying object positions and textures). Specifically, given $N$ parallel environments and a required group size of $G$, we sample $N/G$ independent random seeds. We then duplicate each seed $G$ times via sequence interleaving. This implementation guarantees two critical properties: (1) \textbf{Intra-group consistency}: all $G$ rollouts within a single group are initialized with the exact same physical configuration, providing a perfectly aligned baseline for unbiased relative advantage estimation; (2) \textbf{Inter-group diversity}: different groups receive distinct seeds, preserving the robustness and generalization benefits inherent to domain randomization.

\subsection{World-Model-Generated Interactions}

To examine whether \textbf{T$^2$VLA} can operate beyond direct interaction with standard simulators, we additionally use OpenSora~\cite{zheng2024open} as a learned world model. During policy optimization, OpenSora synthesizes action-conditioned observations for exploratory interactions on LIBERO-Spatial. The resulting OpenVLA-OFT policy is then evaluated in the original LIBERO-Spatial simulator. Thus, OpenSora serves as the source of model-generated interaction observations rather than as an independent evaluation benchmark.

\section{Further Analysis of Confidence-Based Expert Election}
\label{sec:further_confidence_analysis}

\subsection{Generalization across Models and Interaction Environments}

\begin{figure}[htbp]
    \centering
    \includegraphics[width=0.95\linewidth]{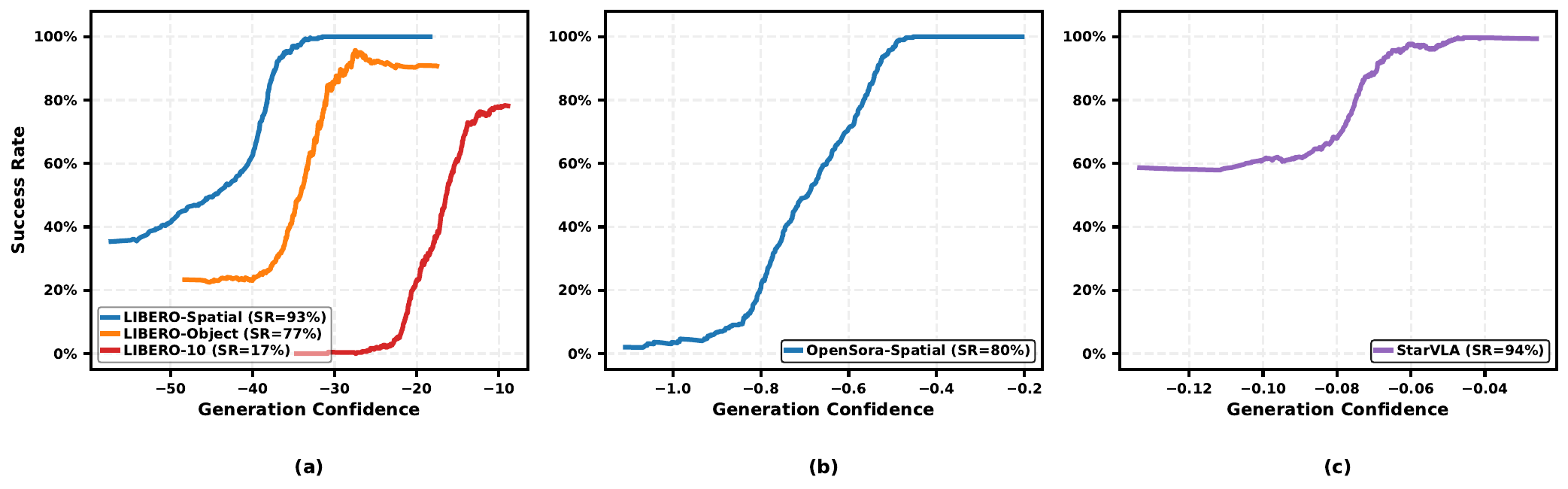}
    \caption{\textbf{Generalization of confidence--success relationships.}
    (a) OpenVLA-OFT on LIBERO-Spatial, LIBERO-Object, and the low-success-rate LIBERO-10 setting.
    (b) OpenVLA-OFT interacting with observations synthesized by an OpenSora world model.
    (c) StarVLA on LIBERO-Spatial.
    Across these discrete-action VLA settings, trajectories with higher generation confidence generally achieve higher task success rates.}
    \label{fig:supp_confidence_generalization}
\end{figure}

In our main manuscript, we observe that trajectories with higher generation confidence are more likely to succeed for discrete-action VLAs. Here, we further investigate whether this empirical relationship persists across different initial policy strengths, interaction environments, and VLA architectures. Figure~\ref{fig:supp_confidence_generalization}(a) compares OpenVLA-OFT across LIBERO-Spatial, LIBERO-Object, and LIBERO-10. In particular, the relationship remains visible on LIBERO-10 even when the initial success rate is only approximately $17\%$, indicating that the confidence ordering is not limited to already strong policies.

We additionally evaluate OpenVLA-OFT using action-conditioned observations synthesized by an OpenSora world model~\cite{zheng2024open}. As shown in Figure~\ref{fig:supp_confidence_generalization}(b), higher-confidence trajectories remain more likely to succeed when evaluated under model-generated observations. Figure~\ref{fig:supp_confidence_generalization}(c) presents the same analysis for StarVLA~\cite{community2026starvla} on LIBERO-Spatial and exhibits a similar trend. These results show that the observed confidence--success relationship persists across the evaluated discrete-action VLA architectures and interaction settings.

\subsection{Confidence-Based Expert Election for Continuous VLAs}

\begin{figure}[htbp]
    \centering
    \includegraphics[width=0.95\linewidth]{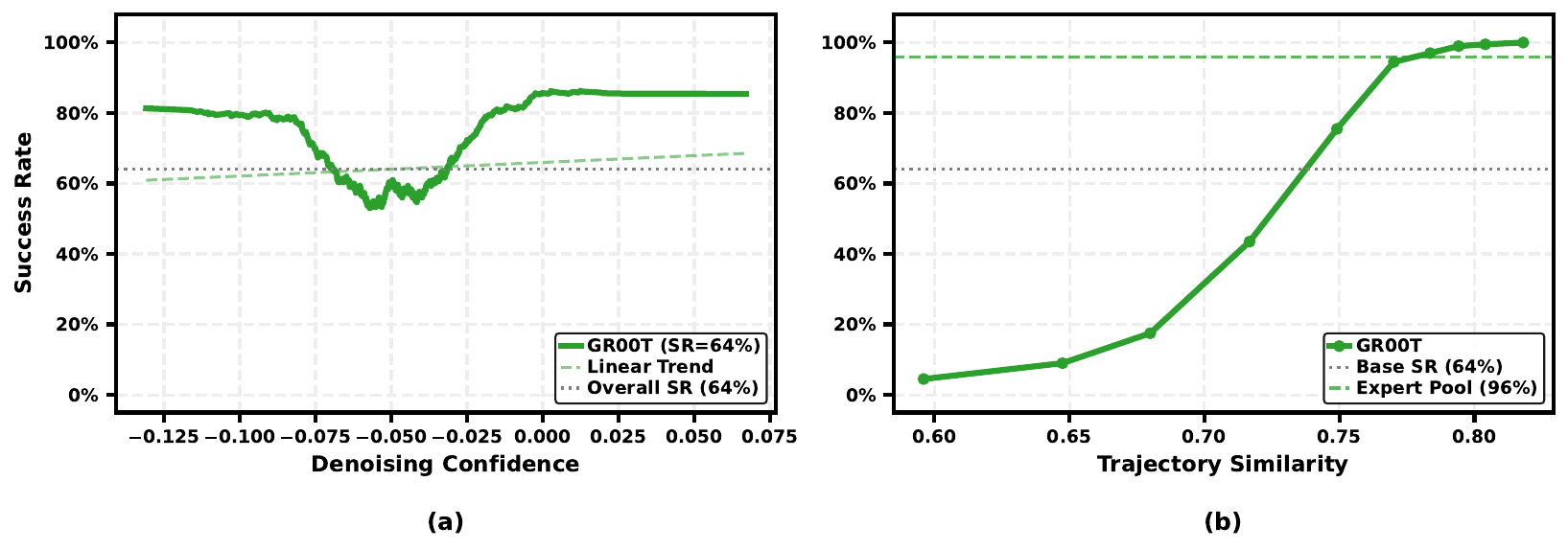}
    \caption{\textbf{Expert election with the continuous-action GR00T policy on LIBERO-Object.}
    (a) Relationship between denoising confidence and task success.
    (b) Relationship between DTW similarity to the elected experts and task success.
    The elected high-confidence experts achieve a $96\%$ success rate, and similarity to these experts provides an informative reward signal.}
    \label{fig:supp_continuous_confidence}
\end{figure}

For continuous-action VLAs, the likelihood formulation differs from the token log-probability used by autoregressive policies. GR00T~\cite{bjorck2025gr00t} estimates confidence using per-step Gaussian likelihoods along its denoising process, which characterize denoising consistency. Figure~\ref{fig:supp_continuous_confidence}(a) presents the relationship between trajectory-level denoising confidence and task success rate.

\textbf{T$^2$VLA} does not directly optimize the policy using the numerical confidence score. Confidence is used to rank exploratory rollouts and elect high-confidence trajectories as pseudo-experts. These elected trajectories subsequently serve as behavioral anchors, and the intrinsic reward for each rollout is computed from its DTW-based trajectory similarity to the selected experts.

Among $2{,}000$ GR00T rollouts on LIBERO-Object, the high-confidence trajectories elected as experts achieve a $96\%$ success rate, compared with an overall rollout success rate of approximately $64\%$. This result shows that confidence-based election yields a reliable expert set in the evaluated setting.

To further assess whether the elected experts provide useful behavioral anchors, Figure~\ref{fig:supp_continuous_confidence}(b) groups the remaining rollouts according to their DTW similarity to these experts. The success rate generally increases with expert similarity, indicating that trajectories more closely aligned with the elected experts are also more likely to complete the task. Together with the high success rate of the elected experts, this result supports the two-stage design of \textbf{T$^2$VLA}: confidence identifies high-quality reference trajectories, and similarity to these references provides an informative intrinsic signal for policy optimization. One limitation is that confidence-based election may omit successful but low-confidence trajectories, motivating future work on likelihood-aware expert selection.

\subsection{Confidence Ordering during Policy Optimization}

\begin{figure}[htbp]
    \centering
    \includegraphics[width=0.9\linewidth]{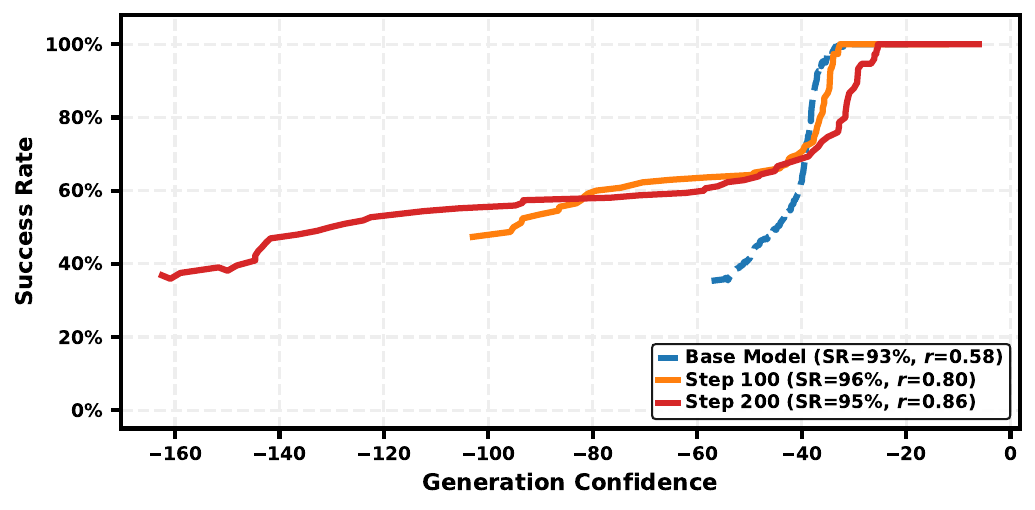}
    \caption{\textbf{Confidence--success relationship during policy optimization.}
    OpenVLA-OFT checkpoints at the initial policy, step 100, and step 200 are evaluated on LIBERO-Spatial. The correlation remains positive and strengthens from $r=0.58$ to $r=0.80$ and $r=0.86$, respectively.}
    \label{fig:supp_confidence_training}
\end{figure}

\textbf{T$^2$VLA} only requires confidence to preserve an ordinal relationship suitable for relative expert ranking, rather than to represent a calibrated probability of physical success. To examine whether this ordering deteriorates during optimization, we evaluate the initial OpenVLA-OFT policy and the checkpoints at steps 100 and 200. As shown in Figure~\ref{fig:supp_confidence_training}, the confidence--success correlation remains positive throughout training and progressively strengthens from $r=0.58$ to $r=0.80$ and $r=0.86$. We therefore do not observe degradation of the confidence ordering or emerging overconfidence within the evaluated training horizon.

\section{More Experimental Results}
\label{sec:more_experiments}

We further evaluate \textbf{T$^2$VLA} with additional VLA architectures and manipulation environments. For continuous-action VLA evaluation, we apply our framework to GR00T~\cite{bjorck2025gr00t} on the LIBERO-Spatial and LIBERO-Object suites. We also evaluate the flow-based $\pi_0$ policy on the RoboCasa~\cite{nasiriany2024robocasa} Close Drawer task, which introduces a different household simulation environment from the LIBERO benchmarks used in the main experiments. In all settings, success rate is measured by executing the optimized policy in the corresponding evaluation environment.

\begin{table}[htbp]
    \centering
    \small
    \caption{\textbf{Results across additional models and environments.} Success rates (\%).}
    \label{tab:supp_additional_settings}
    \begin{tabular*}{\linewidth}{@{\extracolsep{\fill}}ccccc@{}}
        \toprule
        \textbf{Model} & \textbf{Setting} & \textbf{Base} & \textbf{Ours} & \textbf{$\Delta$} \\
        \midrule
        GR00T & LIBERO-Spatial & 41.4 & \textbf{60.9} & \textit{+19.5} \\
        GR00T & LIBERO-Object  & 58.6 & \textbf{98.8} & \textit{+40.2} \\
        $\pi_0$ & RoboCasa Close Drawer & 75.0 & \textbf{87.5} & \textit{+12.5} \\
        \bottomrule
    \end{tabular*}
\end{table}

As shown in Table~\ref{tab:supp_additional_settings}, our method improves GR00T by $19.5$ points on LIBERO-Spatial and $40.2$ points on LIBERO-Object. On RoboCasa Close Drawer, the success rate of $\pi_0$ improves from $75.0\%$ to $87.5\%$. These results provide additional evidence that the proposed expert-election and trajectory-alignment formulation applies across VLA architectures and manipulation environments.

\section{Mathematical Formulations of Advantage Shaping Strategies}
\label{sec:appendix_shaping_math}

In this section, we provide the detailed mathematical definitions for the expert fusion and advantage shaping strategies compared in Table 5 of the main manuscript (Section 4.4). Let $c_{local, l}^*$ denote the confidence score of the current task-conditioned local expert, and $\mathcal{P}_l$ represent the historical expert pool with mean confidence $\bar{c}_{pool}$, maximum $c_{max, l}$, and minimum $c_{min, l}$. The fusion weight $w \in [0, 1]$ determines the contribution of each expert source to the hybrid similarity reward.

\begin{itemize}[leftmargin=*]
    \item \textbf{Static Weighting:} A naive baseline where the current batch and historical pool are weighted equally regardless of their relative quality:
    \begin{equation}
        w = 0.5
    \end{equation}
    
    \item \textbf{Max Routing:} A hard-selection strategy that assigns full weight to the more confident source:
    \begin{equation}
        w = \begin{cases} 1 & \text{if } c_{local}^* \ge \bar{c}_{pool} \\ 0 & \text{otherwise} \end{cases}
    \end{equation}
    
    \item \textbf{Adaptive Margin Fallback (AMF):} A gating mechanism designed to favor on-policy exploration unless the current batch quality drops significantly below the historical average by a fixed margin $m=2.0$:
    \begin{equation}
        w = \mathbb{I}(c_{local}^* \ge \bar{c}_{pool} - m)
    \end{equation}
    
    \item \textbf{Temperature-scaled Sigmoid Gate:} A soft-gating mechanism that provides continuous weight transitions based on the confidence gap, utilizing a temperature parameter $T=2.0$:
    \begin{equation}
        w = \frac{1}{1 + \exp\left(-\frac{c_{local}^* - \bar{c}_{pool}}{T}\right)}
    \end{equation}
    
    \item \textbf{Asymmetric Z-Score Gate:} A distribution-aware scaling method that uses the running mean $\mu_{pool}$ and standard deviation $\sigma_{pool}$ of the expert pool. It applies higher sensitivity to high-confidence discovery:
    \begin{equation}
        w = \sigma(k \cdot Z), \quad Z = \frac{c_{local}^* - \mu_{pool}}{\sigma_{pool} + \epsilon}
    \end{equation}
    where $k=2.5$ for $Z \ge 0$ and $k=5.0$ for $Z < 0$ to aggressively discount suboptimal batches.
    
    \item \textbf{Min-Max Confidence Scaling (Ours):} Our proposed strategy maps the non-stationary log-probabilities into a continuous, strictly bounded scale $[0, 1]$ using the extrema of the expert pool:
    \begin{equation}
        w = \text{clip}\left( \frac{c_{local, l}^* - c_{min, l}}{c_{max, l} - c_{min, l} + \epsilon}, 0, 1 \right)
    \end{equation}
\end{itemize}

As empirically observed in the main manuscript, strategies lacking strictly bounded normalization (e.g., Static Weighting and Max Routing) often lead to abrupt gradient shifts and early policy collapse. Our Min-Max scaling ensures an adaptive and smooth integration of expertise, which is essential for stable self-bootstrapping during test-time reinforcement learning.

\section{Algorithmic Implementation}
\label{sec:algo_implementation}
The optimization of \textbf{T$^2$VLA} is implemented as a self-bootstrapping reinforcement learning process. Instead of relying on external environment rewards, reference behaviors are extracted from exploratory rollouts using intrinsic confidence, which then guide policy updates via alignment-based rewards. The complete procedure is summarized in Algorithm~\ref{alg:dual_expert_grpo}, operating in four sequential phases:

\begin{itemize}[leftmargin=*]
    \item \textbf{Phase 1: Exploration \& Confidence Estimation.} The policy $\pi_\theta$ samples a batch of trajectories $\mathcal{D}_l$. Each trajectory is assigned a length-normalized confidence score $c_i$, defined as the mean log-probability of all generated actions, to ensure a length-agnostic evaluation.
    
    \item \textbf{Phase 2: Dual Expert Bootstrapping.} The trajectory with the highest $c_i$ in the current batch is designated as the \textit{Local Pseudo-Expert}. It is then used to update a \textit{Global Expert Pool} $\mathcal{P}_l$, a priority buffer that retains the top-$K$ historical trajectories to capture high-quality reference behaviors.
    
    \item \textbf{Phase 3: Hybrid Reward Synthesis.} Dynamic Time Warping (DTW) is utilized to compute spatial alignment between rollouts and experts, effectively decoupling spatial geometry from temporal variance. The final reward is a dynamically weighted fusion of local and global similarities, regularized by a KL-divergence penalty against $\pi_{\text{ref}}$.
    
    \item \textbf{Phase 4: Policy Optimization.} Policy parameters $\theta$ are updated via GRPO. By estimating advantages through group-level reward normalization instead of a critic network, the framework directly optimizes the policy to maximize the self-bootstrapped rewards.
\end{itemize}

\begin{algorithm}[h]
\caption{Self-Rewarding VLA Adaptation via Confidence-Driven Dual Experts}
\label{alg:dual_expert_grpo}
\begin{algorithmic}[1] 
\Require Pretrained VLA $\pi_\theta$, reference VLA $\pi_{\text{ref}}$, instruction $l$, batch size $N$, pool capacity $K$
\Statex \textbf{Initialize:} Global Expert Pool $\mathcal{P}_{l} \leftarrow \emptyset$

\vspace{1.5mm}
\Statex \textbf{=== Phase 1: Trajectory Rollout \& Confidence Estimation ===}
\State Sample a batch of $N$ trajectories $\mathcal{D}_l = \{\tau_1, \dots, \tau_N\}$ from current policy $\pi_\theta$ conditioned on instruction $l$
\For{$i = 1$ \textbf{to} $N$}
    \State Compute length-normalized confidence: $c_i \leftarrow \frac{1}{T_i} \sum_{t=1}^{T_i} \log \pi_\theta(a_{i,t} | s_{i,t}, l)$
\EndFor

\vspace{1.5mm}
\Statex \textbf{=== Phase 2: Dual Expert Bootstrapping ===}
\State Identify the task-conditioned local expert: $\tau_{local, l}^* \leftarrow \arg\max_{\tau_i \in \mathcal{D}_l} c_i$
\State Let $c_{local, l}^*$ be the confidence score of $\tau_{local, l}^*$
\State Update global pool: $\mathcal{P}_{l} \leftarrow \mathcal{P}_{l} \cup \{(\tau_{local, l}^*, c_{local, l}^*)\}$
\State Retain only the top-$K$ trajectories in $\mathcal{P}_{l}$ sorted by confidence scores

\vspace{1.5mm}
\Statex \textbf{=== Phase 3: DTW-based Hybrid Similarity Reward ===}
\State Get max ($c_{max, l}$) and min ($c_{min, l}$) confidence scores in the current pool $\mathcal{P}_{l}$
\State Compute dynamic interpolation weight: $w \leftarrow \text{clip}\left(\frac{c_{local, l}^* - c_{min, l}}{c_{max, l} - c_{min, l} + \epsilon}, 0, 1\right)$
\For{$i = 1$ \textbf{to} $N$}
    \State Local alignment: $s_{local} \leftarrow \text{Sim}_{\text{DTW}}(\tau_i, \tau_{local, l}^*)$
    \State Global alignment: $s_{global} \leftarrow \max_{\tau_p \in \mathcal{P}_{l}} \text{Sim}_{\text{DTW}}(\tau_i, \tau_p)$
    \State Assign self-supervised reward: $r_i \leftarrow w \cdot s_{local} + (1 - w) \cdot s_{global} - \beta \sum_{t=1}^{T_i} \mathbb{D}_{\text{KL}} \left( \pi_\theta(\cdot | s_{i,t}, l) \| \pi_{\text{ref}}(\cdot | s_{i,t}, l) \right)$
\EndFor

\vspace{1.5mm}
\Statex \textbf{=== Phase 4: Policy Optimization via GRPO ===}
\State Calculate batch reward mean $\mu(r)$ and standard deviation $\sigma(r)$
\For{$i = 1$ \textbf{to} $N$}
    \State Compute Group-normalized Advantage: $A_i \leftarrow \frac{r_i - \mu(r)}{\sigma(r) + \epsilon}$
\EndFor
\State Update policy parameters $\theta$ by maximizing the GRPO surrogate objective using $A_i$
\end{algorithmic}
\end{algorithm}

\section{Analysis of Raw Confidence-based Optimization}
\label{appendix:confidence_analysis}
\begin{figure}[htbp]
    \centering
    \includegraphics[width=1\textwidth]{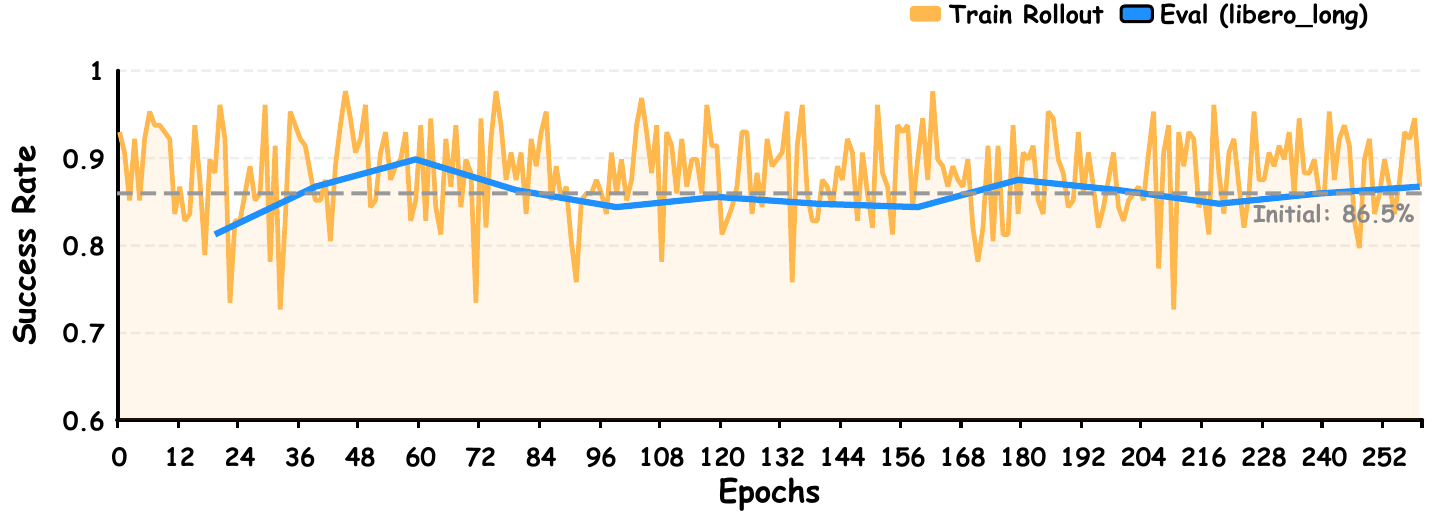}
    \caption{Empirical analysis of pure confidence-based rewards on the \texttt{LIBERO-Long} benchmark. The curves demonstrate that confidence-driven optimization initially boosts the evaluation performance (blue) to around 90\%. However, it eventually plateaus and degrades below the 86.5\% baseline, remaining consistently below the training success rate (orange). This dynamic suggests that a single scalar metric lacks sufficient signal for sustained optimization, highlighting the need for trajectory-level physical rewards.}
    \label{fig:conf_failure}
\end{figure}
Following the empirical observation in our main manuscript (Section 3.1) that generation confidence correlates with task success, we investigate the viability of using raw log-probabilities directly as the reward signal for policy optimization. We conduct an exploratory experiment on the \texttt{LIBERO-Long} benchmark, where the policy is optimized to maximize trajectory-level mean log-probabilities. 

As illustrated in Figure~\ref{fig:conf_failure}, the confidence-driven objective yields an initial performance improvement. Around epoch 60, the evaluation success rate (blue curve) reaches a peak of approximately 90\%, surpassing the 86.5\% baseline. Throughout the optimization process, the training rollout success rate (orange curve) maintains a consistently high level, fluctuating between 85\% and 95\%. However, following the initial peak, the evaluation performance plateaus and eventually degrades to roughly 84\%--85\%. Consequently, the evaluation curve remains consistently below the training curve for the remainder of the training process.

These empirical results suggest that while the scalar confidence metric effectively reflects the overall certainty of the generated sequence, using it as a direct reward introduces optimization instabilities. Specifically, relying purely on raw log-probabilities can lead to the misclassification of valid physical executions. During exploration, the policy may generate multiple trajectories that all successfully complete the task but yield varying confidence scores due to internal model uncertainties. Consequently, the optimization process might penalize perfectly valid, successful trajectories simply because their relative log-probabilities are lower than others within the same batch. This inconsistent reward assignment confuses the policy update process, ultimately causing the evaluation performance to stagnate without further improvement.

This motivates the reward design in \textbf{T$^2$VLA}, which explicitly combines the scalar confidence metric with physical execution trajectories. To prevent valid executions from being erroneously penalized, we restrict the scalar metric to identifying high-quality reference experts. Subsequently, we use DTW to measure the spatial alignment between exploratory rollouts and these selected physical references. This combined formulation provides a trajectory-level reward, ensuring consistent optimization signals for stable policy learning.

\section{Analysis of Learning Dynamics}
\label{sec:learning_dynamics}
\begin{figure}[htbp]
    \centering
    \includegraphics[width=1\textwidth]{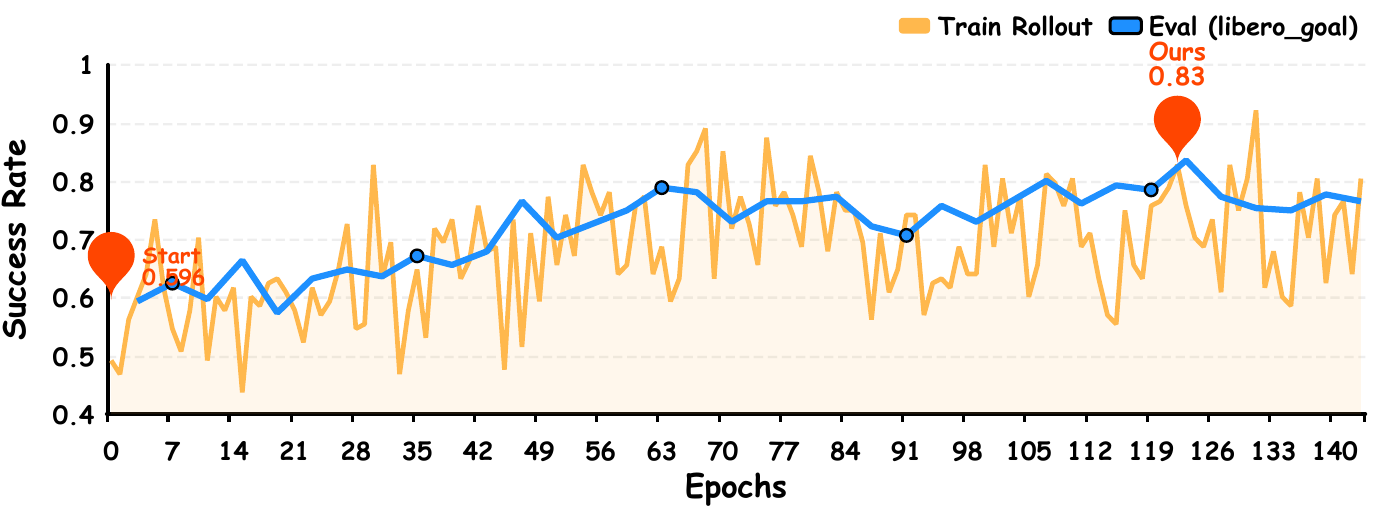}
    \caption{Learning dynamics on the \texttt{LIBERO-Goal} (\texttt{traj1}) benchmark. The plot tracks the success rates of exploratory training rollouts (orange) and periodic evaluations (blue). \textbf{T$^2$VLA} steadily improves the policy from an initial $59.6\%$ to a peak of $83.0\%$ using exclusively intrinsic rewards, with environment signals reserved strictly for monitoring.}
    \label{fig:learning_curve}
\end{figure}
To illustrate the optimization stability of \textbf{T$^2$VLA}, Figure~\ref{fig:learning_curve} presents the learning curves on the \texttt{LIBERO-Goal} benchmark under the 1-shot (\texttt{traj1}) setting, where the baseline policy is fine-tuned on a single demonstration trajectory. This specific model is  selected as a representative case study. Its moderate initial success rate ($59.6\%$) exceeds the minimum bootstrapping threshold, preventing the optimization collapse observed in low-competence regimes. Furthermore, it leaves ample margin for continuous policy refinement, clearly demonstrating the progressive learning dynamics. The optimization process is driven exclusively by the proposed self-bootstrapped intrinsic rewards. While ground-truth environment success rates are recorded concurrently to monitor actual task progression, these external signals are strictly excluded from gradient computation and policy updates.

As depicted in Figure~\ref{fig:learning_curve}, the baseline policy initiates with an evaluation success rate of $59.6\%$ at epoch 0. Over the course of 140 training epochs, both the training rollout performance and the periodic evaluation success rate demonstrate a consistent upward trend. Facilitated by the dual-expert reward formulation and dynamic confidence scaling, the framework systematically refines its execution capabilities based on self-generated experiences.

The evaluation trajectory reveals a rapid initial ascent, reaching approximately $79.0\%$ around epoch 63. Following a brief consolidation phase, the success rate resumes its growth, achieving a peak of $83.0\%$ near epoch 119. This detailed dynamic confirms that the framework effectively extracts viable optimization signals from intrinsic model confidence, securing substantial policy improvement entirely independent of external environmental feedback.

% \begin{figure}[H]
%     \centering
%     \includegraphics[width=\textwidth]{supp_figures/kl_GN_traj1.pdf}
%     \caption{Learning curves and optimization dynamics under the representative LIBERO-Goal (\texttt{traj1}) setting. (a) The orange curve represents the success rate of the exploratory \textit{Train Rollout}, while the blue curve tracks the \textit{Eval} success rate conducted every 4 epochs, improving from $0.596$ to $0.83$. (b) The policy gradient norm maintains a healthy magnitude and decays smoothly, ensuring stable optimization. (c) The policy KL divergence remains strictly bounded, preventing catastrophic forgetting.}
%     \label{fig:learning_curve}
% \end{figure}
% Beyond task performance, analyzing the underlying optimization dynamics further validates the stability of our framework. As shown in Figure~\ref{fig:learning_curve}(b), the policy gradient norm maintains a healthy $\mathcal{O}(1)$ magnitude and decays smoothly upon convergence. This confirms that our bounded reward formulation successfully prevents the vanishing gradient issue inherently caused by raw, unscaled confidence scores. Furthermore, Figure~\ref{fig:learning_curve}(c) demonstrates that the policy KL divergence is strictly bounded within a narrow margin ($0.01 \sim 0.02$). This ensures that \textbf{T$^2$VLA} rigorously adheres to trust-region constraints, enabling continuous self-improvement while completely avoiding catastrophic forgetting of pre-trained physical priors.

\section{Details of Ablation Studies on the Dual Expert Mechanism}
\label{sec:ablation_learning_dynamics}
\begin{figure}[htbp]
    \centering
    \includegraphics[width=\textwidth]{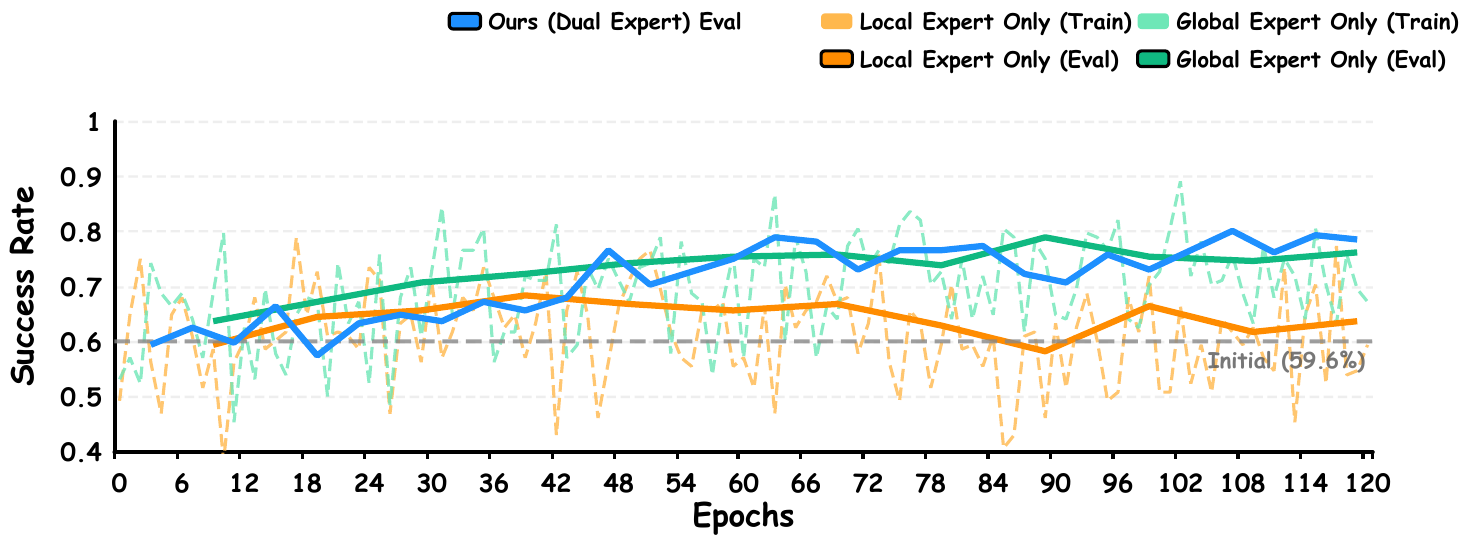}
    \caption{Ablation of the Dual Expert mechanism on the \texttt{LIBERO-Goal} (\texttt{traj1}) benchmark. The plot contrasts the evaluation success rates of the \textit{Local Expert Only} (orange) and \textit{Global Expert Only} (green) configurations against our synergistic \textit{Dual Expert} approach (blue). The initial SFT baseline is denoted by the dashed gray line. The dynamic weighting effectively combines the responsiveness of the local expert with the stability of the global expert to achieve better performance.}
    \label{fig:ablation_curve}
\end{figure}
To provide a detailed analysis of the dual-expert mechanism discussed in our main manuscript (Section 4.3), we align our evaluation with the representative \texttt{LIBERO-Goal} (\texttt{traj1}) setting established in Section~\ref{sec:learning_dynamics}. Figure~\ref{fig:ablation_curve} visualizes the episodic learning dynamics of the isolated and combined expert configurations, starting from the initial $59.6\%$ baseline.

\textbf{Local Expert Configuration.} The policy optimized exclusively via the local expert (orange curve) facilitates active exploration and rapidly captures the latest on-policy feedback. However, because its guidance is strictly bounded by the current sampled batch, the optimization signal remains vulnerable to the variance of immediate exploratory rollouts. This batch-level dependency occasionally introduces suboptimal anchors, causing the evaluation success rate to fluctuate during the training process.

\textbf{Global Expert Configuration.} The global expert pool (green curve) provides a highly stable and consistent optimization signal. Crucially, because this historical pool is constructed entirely from the highest-confidence local experts collected over past iterations, its steady upward trajectory inherently validates the fundamental reliability of the local experts. By maintaining a Top-$K$ memory buffer, this configuration effectively filters out isolated batch-level variance. It ensures that the policy systematically accumulates and retains high-quality behaviors, establishing a robust performance foundation throughout the training process.

\textbf{Dual Expert Synergy.} Our \textbf{T$^2$VLA} framework (blue curve) integrates both components through dynamic confidence scaling. Building upon the solid foundation provided by the global pool, it incorporates the local expert to actively capture highly confident novel behaviors as they emerge. By adaptively balancing the immediate on-policy discoveries with the stable historical references, this synergy maximizes the advantages of both mechanisms, ultimately achieving the highest peak success rate.

\section{Training Hyperparameters}
\label{appendix:hyperparameters}

To facilitate reproducibility, we detail the complete set of hyperparameters used for training our \textbf{T$^2$VLA} framework on the \textbf{LIBERO} benchmark in Table~\ref{tab:hyperparameters}. In this configuration, we build our framework utilizing the OpenVLA-OFT architecture.

\begin{table}[htbp]
    \centering
    \small
    \caption{Hyperparameters for RL with OpenVLA-OFT.}
    \label{tab:hyperparameters}
    % 总宽度依旧死死锁定为 \linewidth 不变
    % >{\centering\arraybackslash}X 保证 X 列（自适应宽度的列）里的文字全部居中
    \begin{tabularx}{\linewidth}{@{} >{\centering\arraybackslash}X c | >{\centering\arraybackslash}X c @{}}
        \toprule
        \textbf{Hyperparameter} & \textbf{Value} & \textbf{Hyperparameter} & \textbf{Value} \\
        \midrule
        
        \multicolumn{2}{c|}{\textit{Training \& Environment}} & \multicolumn{2}{c}{\textit{Optimization \& GRPO}} \\
        \cmidrule(r){1-2} \cmidrule(l){3-4} 
        
        Total training epochs & 300 & Advantage estimator & GRPO \\
        Train batch size & 16 & GRPO group size (samples per task) & 8 \\
        Validation batch size & 128 & PPO update epochs & 1 \\
        Max environment steps & 500 & Actor learning rate & $5 \times 10^{-6}$ (Cosine) \\
        Max prompt length & 256 & Critic learning rate & $1 \times 10^{-5}$ (Constant) \\
        Max response length & 128 & Clip bounds ($\epsilon_{low}$, $\epsilon_{high}$) & (0.2, 0.28) \\
        
        \addlinespace 
        \multicolumn{2}{c|}{\textit{Action Generation}} & Initial KL coefficient & 0.02 \\
        \cmidrule(r){1-2}
        
        Action chunk length & 8 & Target KL coefficient & 0.04 \\
        Action token length & 7 & Entropy coefficient & 0.005 \\
        Rollout temperature & 1.6 & Reward discount rate ($\gamma$) & 1.0 \\
        Top-p sampling & 1.0 & & \\
        
        \bottomrule
    \end{tabularx}
\end{table}

\section{Case Study: Visualizing the Bootstrapping Threshold}
\label{sec:case_study_threshold}

\subsection{Stable Bootstrapping under Strong Initial Priors}
\label{subsec:case_study_success}
\begin{figure}[htbp]
    \centering
    \includegraphics[width=1\textwidth]{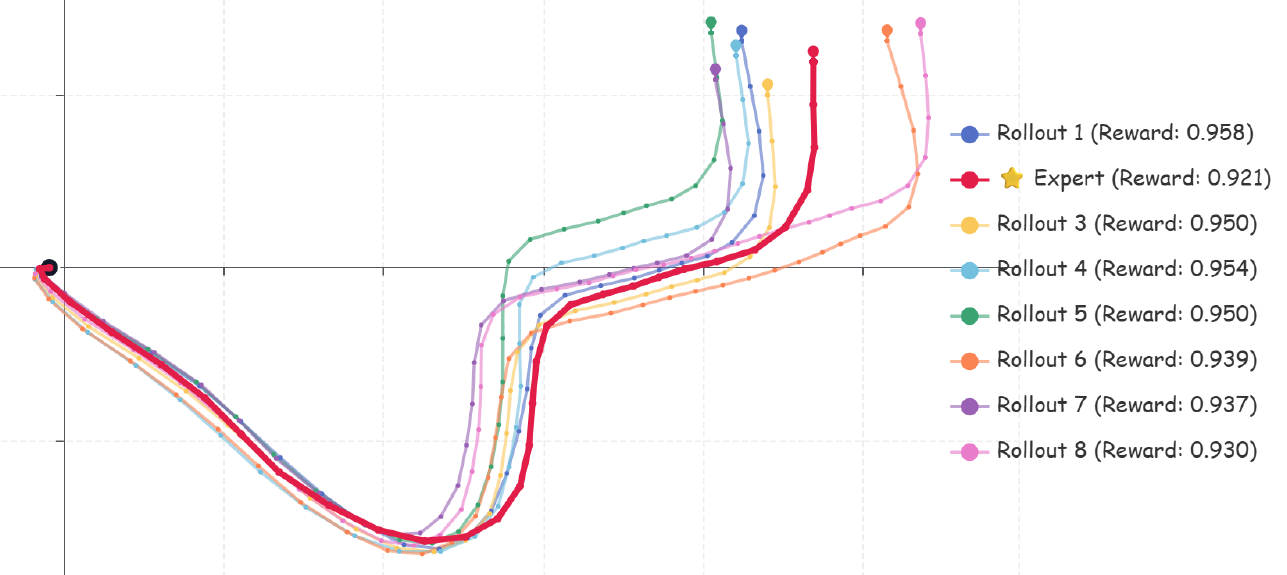}
    \caption{Exploratory rollouts in a high-competence regime (base success rate $86.5\%$). The initial policy ensures all 8 rollouts cover the full task horizon without early termination. The framework selects a structurally complete expert (red line), providing valid anchors for stable policy refinement.}
    \label{fig:case_study_success}
\end{figure}

To visually understand the exploratory dynamics and the mechanism behind the bootstrapping threshold, we implemented an in-situ tracker during training to capture and log the exact micro-batches of exploratory rollouts and their corresponding rewards. We first examine the stable bootstrapping behavior of a highly competent initial policy. Figure~\ref{fig:case_study_success} displays a batch of 8 trajectories generated by a model fine-tuned on full demonstration data, achieving an $86.5\%$ base success rate on the \texttt{LIBERO-10} suite.

As demonstrated by the 2D projection of cumulative delta-actions, all 8 rollouts successfully cover the full spatial horizon of the task. The initial policy maintains consistent directional progression, avoiding premature halting. While the trajectories exhibit local variance due to exploration, they universally adhere to the global task geometry. 

Consequently, the framework evaluates a pool of structurally complete executions. The selected expert (marked with a red star) represents a physically valid, full-length trajectory rather than a degenerate path. This visualization confirms that initial policy competence determines the quality of self-generated trajectories: when the base policy possesses sufficient capability, its exploratory rollouts consistently span the required task horizon. The structural completeness of these trajectories provides the valid anchors necessary for grounded reward formulation, ensuring the stable optimization signals required for effective and continuous policy refinement.

\subsection{Optimization Collapse under Weak Initial Priors}
\label{subsec:case_study_failure}
\begin{figure}[htbp]
    \centering
    \includegraphics[width=1\textwidth]{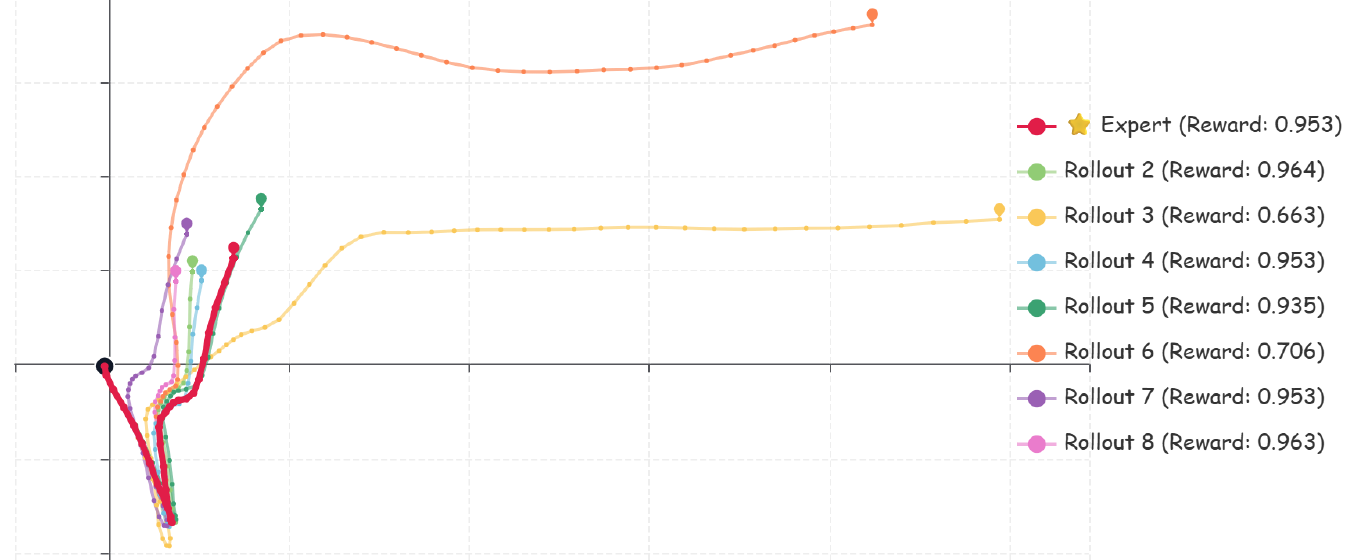}
    \caption{A failure case visualizing the bootstrapping threshold on the \texttt{LIBERO-Long} suite (1-shot setting). The framework evaluates 8 exploratory rollouts generated by a weak initial model and assigns the highest confidence (Reward $= 1.000$) to a spatially truncated trajectory (Expert, red line). Trajectories attempting to explore further (e.g., blue, purple) receive lower scores. Bootstrapping from such incomplete anchors illustrates the optimization challenges in excessively low-competence regimes.}
    \label{fig:case_study_failure}
\end{figure}
To contrast with the stable bootstrapping observed under strong priors, we investigate the optimization collapse under weak priors. As discussed in the main manuscript (Section 4.4), our self-bootstrapping framework exhibits a performance degradation on the \texttt{LIBERO-Long} suite under the extreme 1-shot setting (dropping from $17.3\%$ to $11.0\%$). Figure~\ref{fig:case_study_failure} visualizes a representative batch of 8 trajectories collected at Epoch 0 under a single task instruction. 

For visualization clarity, the cumulative delta-action sequences of the end-effector are projected onto a 2D plane. Given the low initial success rate of the baseline model ($17.3\%$), it typically struggles to handle complex long-horizon scenarios. Consequently, a significant portion of the generated rollouts naturally fails to cover the full task horizon. Relying on the internal log-probabilities, the framework must autonomously nominate an expert from a pool that predominantly consists of incomplete executions. 

As shown in the figure, the trajectory selected as the Expert (marked with a red star, normalized reward $= 1.000$) is notably brief, terminating early in the spatial execution. Conversely, rollouts that attempt to progress further spatially (e.g., Rollout 7 and Rollout 8) are assigned lower scores.

This visualization clarifies the optimization challenges in low-competence regimes. A severely under-trained initial policy exhibits high epistemic uncertainty regarding the later stages of a task. Consequently, the model's internal confidence metric naturally favors shorter paths that remain within familiar, early-stage state distributions, assigning lower log-probabilities to longer trajectories that venture into unfamiliar regions to make spatial progress. 

Bootstrapping from these spatially truncated anchors misaligns the reward signal with actual task completion. The optimization process inadvertently reinforces early-termination behaviors, leading to the observed performance degradation. This case study empirically supports the existence of a minimum initialization threshold: a foundational policy must possess sufficient competence to generate rollouts that adequately span the task horizon, thereby providing valid structural anchors for autonomous self-improvement.

\section{Limitations}
\label{sec:limitations}

\textbf{Omission of Low-Confidence Experts.}
Continuous-action VLAs formulate trajectory likelihood using Gaussian likelihoods along the denoising process. Successful trajectories assigned relatively low likelihoods may be omitted during expert election.

\textbf{Initialization Threshold.}
With weak initial task competence, particularly on long-horizon tasks, successful or reasonably complete trajectories occur less frequently, reducing the number of high-quality behavioral references.

\textbf{Physical Deployment.}
Our experiments are primarily conducted in simulation and with world-model-generated interactions. Testing and adaptation on physical robots remain future work.

\end{document}